\documentclass{article}

\usepackage{geometry}
\PassOptionsToPackage{numbers, sort&compress}{natbib}
\usepackage{natbib}

\usepackage{times}
\usepackage{helvet}
\usepackage{courier}
\usepackage{multirow}

\usepackage[utf8]{inputenc} 
\usepackage[T1]{fontenc}    
\usepackage[colorlinks=true, citecolor=green]{hyperref}       
\usepackage{url}            
\usepackage{booktabs}       

\usepackage{subcaption}

\usepackage{amsmath}
\usepackage{amssymb}
\usepackage{amsthm}
\usepackage{graphicx}
\graphicspath{figs}
\usepackage{mathtools}

\usepackage{tikz}
\usepackage{array}

\usepackage{algorithm}
\usepackage{algpseudocode}
\usepackage{algorithmicx}

\DeclareMathOperator*{\E}{\mathbb{E}}

\DeclareMathOperator{\Fc}{\mathcal{F}}

\DeclareMathOperator{\Lc}{\mathcal{L}}
\DeclareMathOperator{\Ic}{\mathcal{I}}
\DeclareMathOperator{\R}{\mathbb{R}}

\newcommand{\abs}[1]{\left\lvert#1\right\rvert}

\DeclareMathOperator*{\argmin}{\arg\!\min}

\newtheorem{theorem}{Theorem}
\newtheorem{corollary}{Corollary}[theorem]
\newtheorem{lemma}[theorem]{Lemma}
\newtheorem{definition}{Definition}

\newcommand*\lxor{\oplus}

\usepackage{fullpage}
\usepackage{authblk}

\title{Purifying Interaction Effects with the Functional ANOVA: \\An Efficient Algorithm for Recovering Identifiable Additive Models}

\author[1,2]{Benjamin Lengerich\thanks{blengeri@cs.cmu.edu}}
\author[3]{Sarah Tan}
\author[2,4]{Chun-Hao Chang}
\author[3]{Giles Hooker}
\author[2]{Rich Caruana\thanks{rcaruana@microsoft.com}}
\affil[1]{\emph{Carnegie Mellon University}}
\affil[2]{\emph{Microsoft Research}}
\affil[3]{\emph{Cornell University}}
\affil[4]{\emph{University of Toronto}}

\begin{document}
\maketitle

\begin{abstract}
Models which estimate main effects of individual variables alongside interaction effects have an identifiability challenge: effects can be freely moved between main effects and interaction effects without changing the model prediction. 
This is a critical problem for interpretability because it permits ``contradictory" models to represent the same function. 
To solve this problem, we propose \emph{pure interaction effects}: variance in the outcome which cannot be represented by any smaller subset of features. 
This definition has an equivalence with the Functional ANOVA decomposition. 
To compute this decomposition, we present a fast, exact algorithm that transforms any piecewise-constant function (such as a tree-based model) into a purified, canonical representation. 
We apply this algorithm to Generalized Additive Models with interactions trained on several datasets and show large disparity, including contradictions, between the effects before and after purification. 
These results underscore the need to specify data distributions and ensure identifiability before interpreting model parameters.
\end{abstract}

\section{Motivation}
\label{sec:intro}
\newcommand{\matscale}{0.85}

\begin{figure}[ht]
    \small
    \centering
    \begin{tabular}{m{0.5cm} >{\centering\arraybackslash} m{0.5cm} >{\centering\arraybackslash} m{1.2cm} >{\centering\arraybackslash} m{1.2cm} >{\centering\arraybackslash} m{2.5cm}}
        & $~f_0$ & $~~~~~~~~~f_1$ & $~~~~~~~~~f_2$ & $~~~~~~~f_3$ \\
        \toprule
        \vspace{-10pt}(a) & \vspace{-10pt}$0.25$ & \vspace{-10pt}\begin{tikzpicture}[thick,scale=\matscale, every node/.style={transform shape}]
                \node [draw, fill=blue!20, thick, shape=rectangle, minimum width=1cm, minimum height=1cm, anchor=center] at (0,0) {-0.25};
                \node [draw, fill=red!20, thick, shape=rectangle, minimum width=1cm, minimum height=1cm, anchor=center] at (0,1) {+0.25};
                \node (x1) at (-1, 0.5) {$X_1$};
                \node (x11) at (-0.7, 1) {\small 1};
                \node (x10) at (-0.7, 0) {\small 0};
            \end{tikzpicture} & \vspace{-10pt}\begin{tikzpicture}[thick,scale=\matscale, every node/.style={transform shape}]
                \node [draw, thick, shape=rectangle, minimum width=1cm, minimum height=1cm, fill=blue!20, anchor=center] at (0,0) {-0.25};
                \node [draw, thick, shape=rectangle, minimum width=1cm, minimum height=1cm, fill=red!20, anchor=center] at (0,1) {+0.25};
                \node (x1) at (-1, 0.5) {$X_2$};
                \node (x11) at (-0.7, 1) {\small 1};
                \node (x10) at (-0.7, 0) {\small 0};
            \end{tikzpicture} & \vskip 5pt
        \begin{tikzpicture}[thick,scale=\matscale, every node/.style={transform shape}]
                \node [draw, thick, shape=rectangle, minimum width=1cm, minimum height=1cm, anchor=center] at (0,0) {0};
                \node [draw, thick, shape=rectangle, minimum width=1cm, minimum height=1cm, anchor=center] at (0,1) {0};
                \node [draw, thick, shape=rectangle, minimum width=1cm, minimum height=1cm, anchor=center] at (1,0) {0};
                \node [draw, thick, shape=rectangle, minimum width=1cm, minimum height=1cm, fill=blue!20, anchor=center] at (1,1) {-1};
                \node (x1) at (-1, 0.5) {$X_1$};
                \node (x11) at (-0.7, 1) {\small 1};
                \node (x10) at (-0.7, 0) {\small 0};
                \node (x2) at (0.5, -1) {$X_2$};
                \node (x21) at (1, -0.7) {\small 1};
                \node (x20) at (0, -0.7) {\small 0};
            \end{tikzpicture}
            \\
        \vspace{-10pt}(b) & \vspace{-10pt}$-0.75$ & \vspace{-10pt}\begin{tikzpicture}[thick,scale=\matscale, every node/.style={transform shape}]
                \node [draw, thick, shape=rectangle, minimum width=1cm, minimum height=1cm, fill=red!20, anchor=center] at (0,0) {+0.25};
                \node [draw, thick, shape=rectangle, minimum width=1cm, minimum height=1cm, fill=blue!20, anchor=center] at (0,1) {-0.25};
                \node (x1) at (-1, 0.5) {$X_1$};
                \node (x11) at (-0.7, 1) {\small 1};
                \node (x10) at (-0.7, 0) {\small 0};
            \end{tikzpicture} & \vspace{-10pt}\begin{tikzpicture}[thick,scale=\matscale, every node/.style={transform shape}]
                \node [draw, thick, shape=rectangle, minimum width=1cm, minimum height=1cm, fill=red!20, anchor=center] at (0,0) {+0.25};
                \node [draw, thick, shape=rectangle, minimum width=1cm, minimum height=1cm, fill=blue!20, anchor=center] at (0,1) {-0.25};
                \node (x1) at (-1, 0.5) {$X_2$};
                \node (x11) at (-0.7, 1) {\small 1};
                \node (x10) at (-0.7, 0) {\small 0};
            \end{tikzpicture} & \vskip 5pt \begin{tikzpicture}[thick,scale=\matscale, every node/.style={transform shape}]
                \node [draw, thick, shape=rectangle, minimum width=1cm, minimum height=1cm, anchor=center] at (0,0) {0};
                \node [draw, thick, shape=rectangle, minimum width=1cm, minimum height=1cm, fill=red!20, anchor=center] at (0,1) {+1};
                \node [draw, thick, shape=rectangle, minimum width=1cm, minimum height=1cm, fill=red!20, anchor=center] at (1,0) {+1};
                \node [draw, thick, shape=rectangle, minimum width=1cm, minimum height=1cm, fill=red!20, anchor=center] at (1,1) {+1};
                \node (x1) at (-1, 0.5) {$X_1$};
                \node (x11) at (-0.7, 1) {\small 1};
                \node (x10) at (-0.7, 0) {\small 0};
                \node (x2) at (0.5, -1) {$X_2$};
                \node (x21) at (1, -0.7) {\small 1};
                \node (x20) at (0, -0.7) {\small 0};
            \end{tikzpicture} \\
        \vspace{-10pt}(c) & \vspace{-10pt}$-0.25$ & \vspace{-10pt}\begin{tikzpicture}[thick,scale=\matscale, every node/.style={transform shape}]
                \node [draw, thick, shape=rectangle, minimum width=1cm, minimum height=1cm, anchor=center] at (0,0) {0};
                \node [draw, thick, shape=rectangle, minimum width=1cm, minimum height=1cm, anchor=center] at (0,1) {0};
                \node (x1) at (-1, 0.5) {$X_1$};
                \node (x11) at (-0.7, 1) {\small 1};
                \node (x10) at (-0.7, 0) {\small 0};
            \end{tikzpicture} & \vspace{-10pt}\begin{tikzpicture}[thick,scale=\matscale, every node/.style={transform shape}]
                \node [draw, thick, shape=rectangle, minimum width=1cm, minimum height=1cm, anchor=center] at (0,0) {0};
                \node [draw, thick, shape=rectangle, minimum width=1cm, minimum height=1cm, anchor=center] at (0,1) {0};
                \node (x1) at (-1, 0.5) {$X_2$};
                \node (x11) at (-0.7, 1) {\small 1};
                \node (x10) at (-0.7, 0) {\small 0};
            \end{tikzpicture} & \vskip 5pt \begin{tikzpicture}[thick,scale=\matscale, every node/.style={transform shape}]
                \node [draw, thick, shape=rectangle, minimum width=1cm, minimum height=1cm, fill=red!20, anchor=center] at (0,1) {\small +0.5};
                \node [draw, thick, shape=rectangle, minimum width=1cm, minimum height=1cm, anchor=center] at (0,0) {0};
                \node [draw, thick, shape=rectangle, minimum width=1cm, minimum height=1cm, anchor=center] at (1,1) {0};
                \node [draw, thick, shape=rectangle, minimum width=1cm, minimum height=1cm, fill=red!20, anchor=center] at (1,0) {\small +0.5};
                \node (x1) at (-1, 0.5) {$X_1$};
                \node (x11) at (-0.7, 1) {\small 1};
                \node (x10) at (-0.7, 0) {\small 0};
                \node (x2) at (0.5, -1) {$X_2$};
                \node (x21) at (1, -0.7) {\small 1};
                \node (x20) at (0, -0.7) {\small 0};
            \end{tikzpicture} \\
            \vspace{-10pt}(d) & \vspace{-10pt}$0$ & \vspace{-10pt} \begin{tikzpicture}[thick,scale=\matscale, every node/.style={transform shape}]
                \node [draw, thick, shape=rectangle, minimum width=1cm, minimum height=1cm, anchor=center] at (0,0) {0};
                \node [draw, thick, shape=rectangle, minimum width=1cm, minimum height=1cm, anchor=center] at (0,1) {0};
                \node (x1) at (-1, 0.5) {$X_1$};
                \node (x11) at (-0.7, 1) {\small 1};
                \node (x10) at (-0.7, 0) {\small 0};
            \end{tikzpicture} & \vspace{-10pt} \begin{tikzpicture}[thick,scale=\matscale, every node/.style={transform shape}]
                \node [draw, thick, shape=rectangle, minimum width=1cm, minimum height=1cm, anchor=center] at (0,0) {0};
                \node [draw, thick, shape=rectangle, minimum width=1cm, minimum height=1cm, anchor=center] at (0,1) {0};
                \node (x1) at (-1, 0.5) {$X_2$};
                \node (x11) at (-0.7, 1) {\small 1};
                \node (x10) at (-0.7, 0) {\small 0};
            \end{tikzpicture} & \vskip 5pt \begin{tikzpicture}[thick,scale=\matscale, every node/.style={transform shape}]
                \node [draw, thick, shape=rectangle, minimum width=1cm, minimum height=1cm, fill=red!20, anchor=center] at (0,1) {\small +0.25};
                \node [draw, thick, shape=rectangle, minimum width=1cm, minimum height=1cm, fill=blue!20, anchor=center] at (0,0) {-0.25};
                \node [draw, thick, shape=rectangle, minimum width=1cm, minimum height=1cm, fill=blue!20, anchor=center] at (1,1) {-0.25};
                \node [draw, thick, shape=rectangle, minimum width=1cm, minimum height=1cm, fill=red!20, anchor=center] at (1,0) {\small +0.25};
                \node (x1) at (-1, 0.5) {$X_1$};
                \node (x11) at (-0.7, 1) {\small 1};
                \node (x10) at (-0.7, 0) {\small 0};
                \node (x2) at (0.5, -1) {$X_2$};
                \node (x21) at (1, -0.7) {\small 1};
                \node (x20) at (0, -0.7) {\small 0};
            \end{tikzpicture}
    \end{tabular}
    \caption{Four realizations of Eq.~\eqref{eq:intx_model} on Boolean variables $X_1$ and $X_2$. 
    In each row, we have an overall intercept $f_0$, main effects $f_1$ and $f_2$, and an interaction effect $f_3$. 
    Red indicates a positive and blue a negative effect. 
    While the four models appear to be different and to yield \emph{contradictory} interpretations, \emph{all four models represent the same function and produce identical outputs}. The fourth model (d) is the purified canonical form returned by our algorithm.
    \label{fig:xor}
    }
\end{figure}
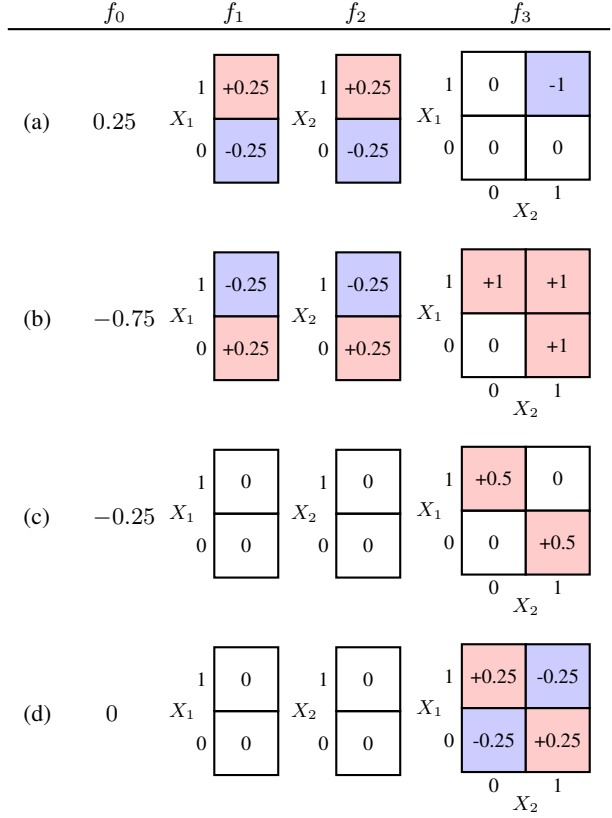

An important question in data analysis is whether two variables act in concert to affect an outcome. 
This question is often approached by estimating an additive model with interactions of the form:
\begin{equation}
    Y \approx f_0 + f_1(X_1) + f_2(X_2) + f_3(X_1, X_2) \label{eq:intx_model}
\end{equation}
and then examining $f_1, f_2, f_3$~\citep{neter1974applied,hastie1990generalized}. 
But this unconstrained additive model has fundamental flaws. We examine two common forms of this model and show that both have problems of identifiability and interpretability.

\subsection{Interactions between Boolean Variables}
\label{sec:intro:boolean}

First, let us consider the simple case of Boolean variables $X_1$ and $X_2$ that take on values $\{0, 1\}$. 
As depicted in Fig.~\ref{fig:xor}, we can represent the additive model with interaction by 3 tables and an intercept. 
The tables represent the main effect of $X_1$, the main effect of $X_2$, and the effect of the interaction between $X_1$ and $X_2$. 
As shown in Fig.~\ref{fig:xor}, we can realize different bitwise operations between $X_1$ and $X_2$ through different values in the interaction table.

Na{\"i}vely, we may believe that the bitwise operations $AND$ (Fig.~\ref{fig:xor}a) and $OR$ (Fig.~\ref{fig:xor}b) represent distinct forms of interaction effect. 
However, we can equivalently write the $AND$ operation as $X_1\land X_2 = -0.25(X_1\lxor X_2) + 0.5(X_1-0.5) + 0.5(X_2-0.5) + 0.25$\footnote{$\lxor$ represents the centered $XOR$ depicted in Fig.~\ref{fig:xor}d.}. 
Similarly, we can write the $OR$ operation as $X_1\lor X_2 = 0.25 (X_1\lxor X_2) + 0.5(X_1-0.5) + 0.5(X_2-0.5) + 0.75$. 
These equivalences make it clear that the interaction effect of $AND$ is identical to the interaction effect of $OR$, and both interactions are actually $XOR$ modified with main effects. 
Thus, the four additive models depicted in Fig.~\ref{fig:xor} are \emph{identical} in their outputs, but generate \emph{contradictory} interpretations. 
Since the main reason for using additive models is to understand the impact of variables and their interactions on the outcome \citep{hastie2017generalized}, this representational degeneracy is problematic.

In this paper, we define interaction effects as variance which \emph{cannot} be explained by main effects. 
Because both $AND$ and $OR$ are the $XOR$ modified with main effects, our definition implies that additive models with interaction effects can always be purified to a weighted $XOR$ interaction. 
This preference has connections to the effect coding representation of inputs \citep{bech2005effects}, discussed in Section~\ref{sec:alg:effect}.

\subsection{Multiplicative Model}
\label{sec:intro:mult}

\begin{figure}[t]
    \centering
    \begin{subfigure}[t]{0.45\textwidth}
        \centering
        \includegraphics[width=0.9\columnwidth]{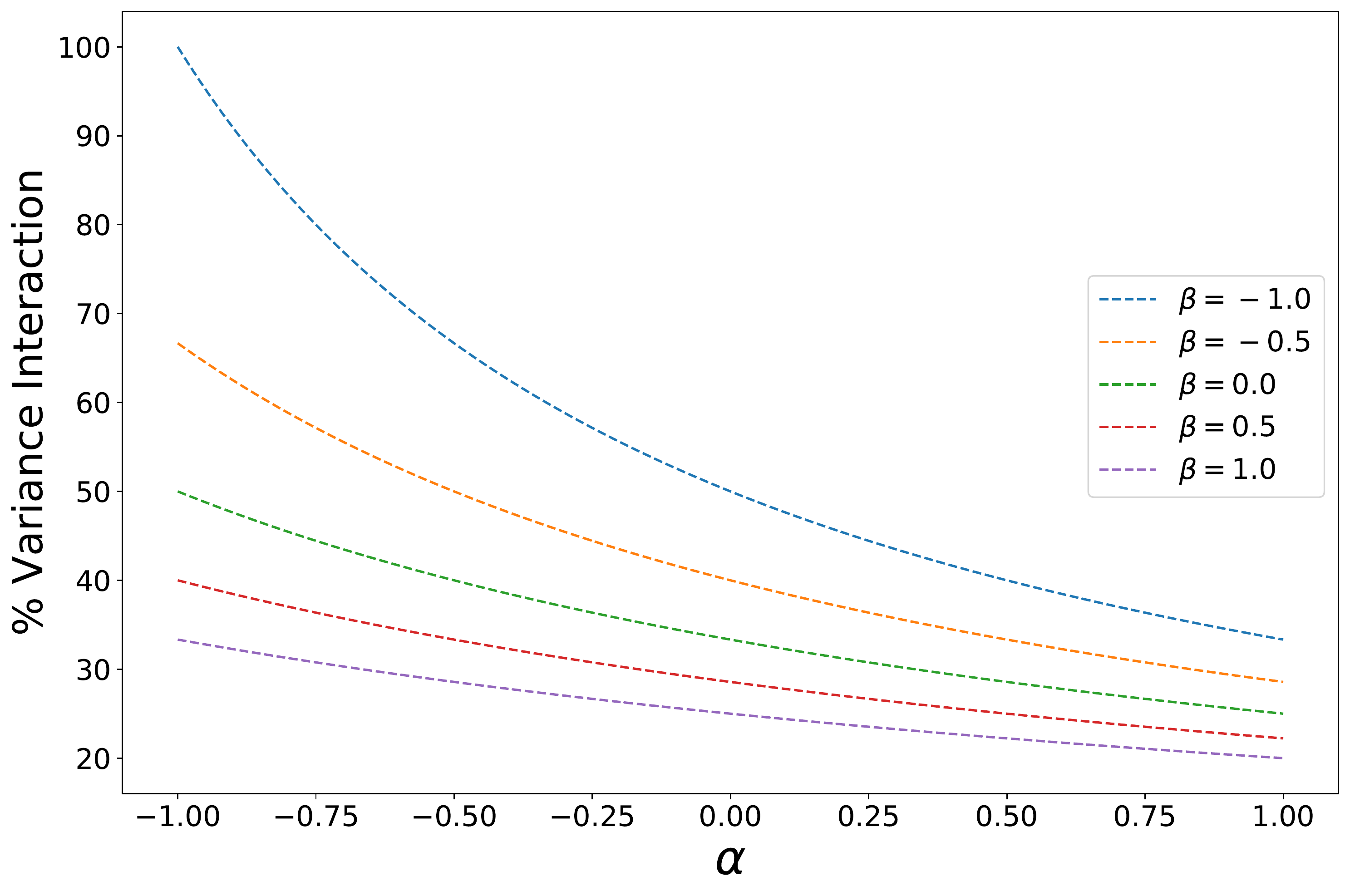}
        \caption{Strength of the interaction effect {\bf implied} by different parameter choices for model \eqref{eq:mult_model_gen}. 
        The vertical axis is the proportion of variance explained by the interaction effect for IID $X_1, X_2\sim N(0, 1)$. 
        In all cases, $a=0$ and $b=c=d=1$, but choice of $\alpha$ and $\beta$ values changes the model's interpretation. 
        In an extreme case, $\alpha=\beta=-1$  makes the main effects disappear entirely. Recall that all of these represent the same function and make the same predictions, only the \emph{interpretations} vary from form to form.}
        \label{fig:mult_intuition_alpha}
    \end{subfigure}
    ~
    \begin{subfigure}[t]{0.45\textwidth}
        \centering
        \includegraphics[width=\columnwidth]{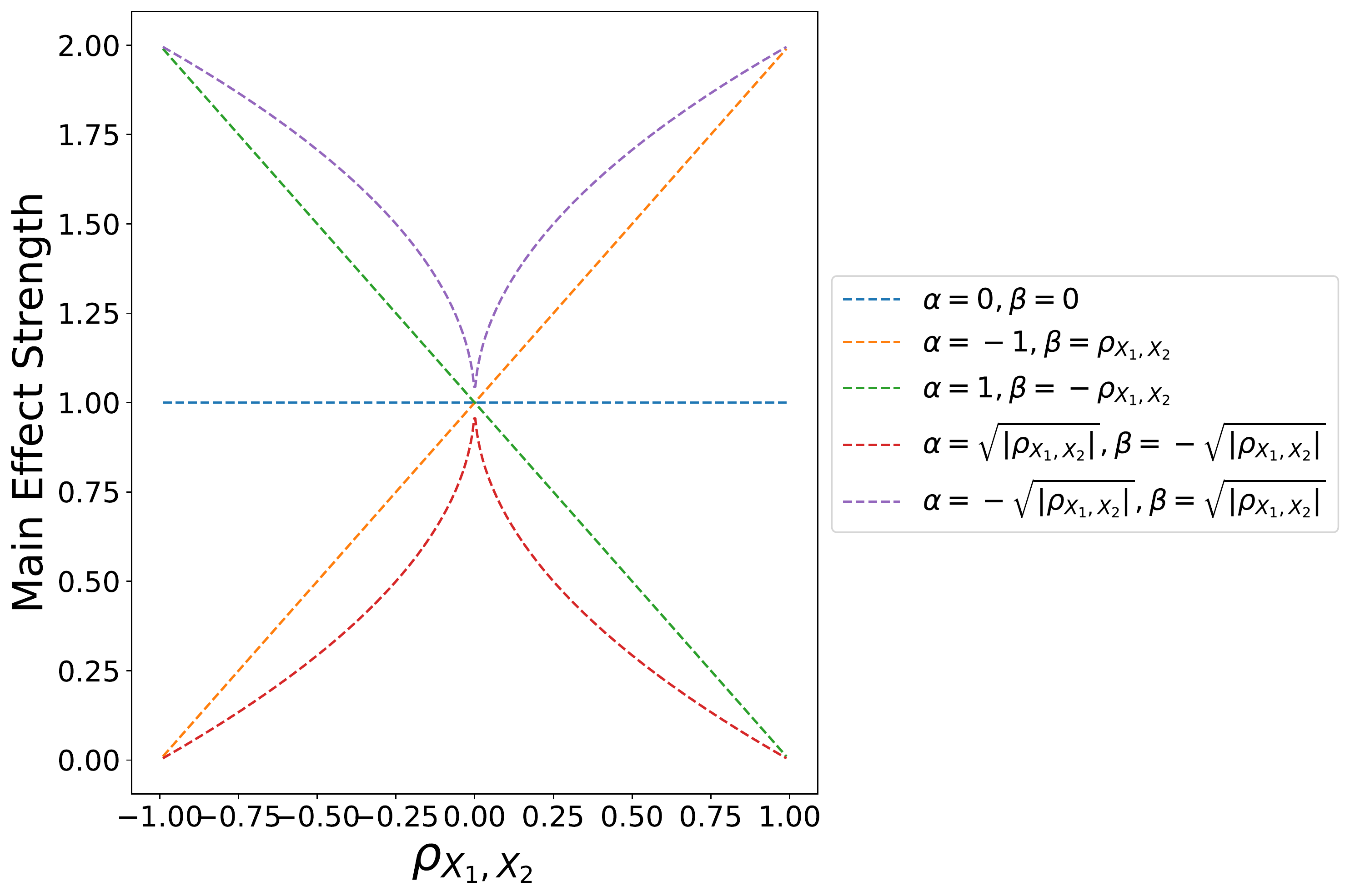}
        \caption{Strength of the main effect of $X_1$ implied by the coefficients of the multiplication model \eqref{eq:mult_model_gen} for various settings of $\alpha, \beta$. 
        For $X_1$, $X_2$ with $\E[X_1] = \E[X_2] = 0$, setting $\alpha=\beta=0$ mean-centers the main effects, but only mean-centers the interaction effect if $X_1$ and $X_2$ are uncorrelated. 
        In contrast, the four settings with $\alpha\beta=-\rho_{X_1,X_2}$ also mean-center $(X_1-\alpha)(X_2-\beta)$. 
        Strength of the main effect of $X_1$ is measured as $b + d\beta$. 
        In all cases, $a=0$ and $b=c=d=1$.
        }
        \label{fig:mult_intuition_centered}
    \end{subfigure}
    \caption{Interpretations of the multiplicative model \eqref{eq:mult_model_gen} change for different settings of $\alpha$ and $\beta$. \label{fig:mult_intuition}}
\end{figure}

Now let us consider the case where $X_1$ and $X_2$ are continuous. 
A common interaction model in statistics \citep{hastie1990generalized} is the linear model augmented with multiplicative features:
\begin{equation}
    Y \approx a + bX_1 + cX_2 + dX_1X_2 \label{eq:mult_model}.
\end{equation}
\noindent
While this model is \emph{identifiable}, the coefficients, unfortunately, are not necessarily \emph{meaningful}. 
For any $\alpha,\beta$, the following model is equivalent to \eqref{eq:mult_model}:
\begin{align}
    Y &\approx (a - d\alpha\beta) + (b+d\beta)X_1 + (c+d\alpha)X_2 \nonumber \\
    &+ (d)(X_1 - \alpha)(X_2 - \beta) \label{eq:mult_model_gen},
\end{align}
The algebraic form of \eqref{eq:mult_model} and its identifiability are a consequence of choosing $\alpha=\beta=0$ in \eqref{eq:mult_model_gen}; however for \emph{any} $\alpha$ and $\beta$, forms \eqref{eq:mult_model} and \eqref{eq:mult_model_gen} make the same predictions. 
As shown in Fig.~\ref{fig:mult_intuition_alpha}, changing the values of $\alpha$ and $\beta$ changes the interpretations of $a, b, c, d$. 

This challenge of selecting $\alpha, \beta$ is not overcome by simply mean-centering all of the effects. 
We can center $X_1, X_2$ such that $\E[X_1]=0, \E[X_2]=0$, which results in $\E[(X_1-\alpha)(X_2 - \beta)] = \rho_{X_1,X_2} + \alpha\beta$. 
Thus, if we would want the intercept $a$ to represent $\E[Y]$, we must select $\alpha,\beta$ such that $\alpha\beta = -\rho_{X_1,X_2}$. 
This selection process has a degree of freedom. 
As shown in Fig.~\ref{fig:mult_intuition_centered}, different choices of  $\alpha, \beta$ lead to very different conclusions about the strengths of the main effects. 
Thus, we need rules governing the selection of values for $\alpha,\beta$. 
In this paper, we propose to follow the functional ANOVA decomposition, which implicitly sets $\alpha,\beta$ such that the variance of interaction terms is minimized. 

\subsection{Contributions}
\label{sec:intro:contributions}

In this paper, our major contributions are threefold:
\begin{itemize}
    \item We study the problem of non-identifiability of additive models with interactions, and show that the functional ANOVA decomposition repairs this flaw. We argue that the functional ANOVA decomposition should be preferred to other representations of interaction. 
    \item We propose a fast, exact algorithm to recover the functional ANOVA decomposition from piece-wise constant functions such as tree-based models. 
    \item We show that na{\"i}ve inspection of popular models for training GAMs with interactions can produce contradictory
    conclusions on real datasets. These contradictions are corrected by our purification algorithm.
\end{itemize}

\section{Related Work}
\label{sec:related}

Generalized additive models (GAMs) have long been used to model individual features flexibly \citep{hastie1990generalized}, using functional forms such as splines, trees, wavelets, etc. \citep{eilers1996flexible,lou2012intelligible, wand2011penalized}. GAMs are claimed to be interpretable \citep{hastie1990generalized,caruana2015intelligible} and have been leveraged for interpretability tasks, such as identifying unexpected relationships between features and predictions. For example, using GAMs, \cite{caruana2015intelligible} found an unexpected relationship between having asthma and decreased likelihood of pneumonia mortality in a medical records dataset.

While vanilla GAMs describe nonlinear relationships between each feature and the label, interactions are sometimes added to further capture relationships between multiple features and the label \citep{coull2001simple,lou2013accurate}. However, the resulting models are overparametrized if the parameters are not regularized or constrained \citep{marascuilo1970appropriate,rosnow1989definition,terbeck1998interactions,green1999overparameterized, davies2012interactions} -- the interactions can hence be non-identifiable and non-unique. 
The interpretability of GAMs may be misleading if the relationship between a feature and the prediction changes after adding an interaction with that feature to the model. To address this, constraints such as the ``sum-to-zero'' restriction \citep{hastie1990generalized} -- where parameters are constrained to sum to zero -- or effect coding \citep{bech2005effects} -- a certain type of one-hot-encoding with fewer degrees of freedom -- have been proposed. 

The functional ANOVA decomposition, which we study in this paper, also addresses this issue using ``integrate-to-zero'' restrictions \citep{hooker2007generalized}. Interestingly, the functional ANOVA has also been used to isolate effects of individual features in settings where many features are changing at a time, such as in hyperparameter tuning \citep{pmlr-v32-hutter14}. However the connection between the functional ANOVA and isolating effects of individual features in interactions has not been studied.

Our definition of interactions as variance which cannot be explained by main effects is similar to \cite{yu2019reluctant}'s ``reluctance'' principle: that a main effect should be preferred over an interaction if both have similar prediction performance. Our definition of interactions is also related to Sobol indices \citep{sobol2001global}, that measure how important a feature or interaction is in terms of the amount of prediction variance explained. 

Due to the computational cost of finding interactions, the search space of possible interactions is typically restricted. For example, the popular hierarchy restriction \citep{bien2013lasso} only considers a potential interaction if its component features are already present in the model as main effects. We briefly mention a few interaction detection methods, and refer the reader to a recent review by \cite{bien2013lasso} for more. They can be roughly divided into two types: hypothesis-testing for interactions \citep{sperlich2002nonparametric}, or model-based methods \cite{tsang2017detecting,purushotham2014factorized}, including methods that use tree-based models to detect interactions \citep{sorokina2008detecting,du2019interaction}. 

Instead of restricting the model class for estimation, our proposed method of purifying interactions is designed to be applied post-hoc after estimation. 
Other post-hoc optimizers include decision tree pruning \citep{mingers1989empirical}, which aims to remove spurious interactions, 
and local models to approximate large models \citep{ribeiro2016should,lengerich2019learning}.

In this paper we focus on tree-based models. In contrast to other recent works on generating post-hoc explanations from tree-based models, such as feature importance, rules, etc. \citep{devlin2019disentangled,pmlr-v84-hara18a,deng2019interpreting}, in this work we focus on defining what purified interactions for tree-based models look like. 

\section{Functional ANOVA}
\label{sec:fanova}

The Functional ANOVA (fANOVA) \citep{hoeffding1948central,stone1994use,huang1998projection,cuevas2004anova,hooker2007generalized} seeks to decompose a function $F(X)$ into:
\begin{equation}
    F(X) = f_0 + \sum_{i=1}^df_i(X_i) + \sum_{i\neq j}f_{ij}(X_i, X_j) + \dots,
\end{equation}
where $X = (X_1,\ldots,X_d)$. 
By the uniqueness of fANOVA under non-degenerate feature distributions \citep{chastaing2012}, this set of functions uniquely defines an orthogonal decomposition of $F$ with minimum variance in higher-order functions. 
From this decomposition, we can uniquely define interaction effects.

\subsection{fANOVA for Continuous Functions}
\label{sec:fanova:continuous}

Given a density $w(X)$ and $\Fc^u \subset \Lc^2(\R^u)$ the family of allowable functions for variable set $u$, the weighted fANOVA \citep{hooker2004diagnostics,hooker2007generalized} seeks:
\begin{subequations}
\begin{align}
    \{f_u(X_u)|u \subseteq [d] \} = \argmin_{\{g_u \in \Fc^u\}_{u\in [d]}} \int \Big(\sum_{u \subseteq [d]}g_u(X_u) - F(X) \Big)^2w(X)dX, 
\end{align}
where $[d]$ indicates the power set of $d$ features, such that
\begin{equation}
    \forall~v \subseteq~u,\quad \int f_u(X_u)g_v(X_v)w(X)dX = 0 \quad~\forall~g_v,
    \label{eq:orthogonality}
\end{equation}
i.e., each member $f_u$ is orthogonal to the members which operate on a subset of the variables in $u$. 
By Lemma 4.1 of \cite{hooker2007generalized}, these orthogonality conditions are equivalent to the integral conditions
\begin{equation}
    \forall~u \subseteq [d], \forall~i~\in~u, \quad \int f_u(X_u)w(X)dX_i dX_{-u} = 0
    \label{eq:integral_conditions}.
\end{equation}
\end{subequations}
where the subscript $-u$ indicates the set of variables not in $u$. 
Thus, we seek a set of functions $f_u$ which jointly satisfy \eqref{eq:integral_conditions} with respect to a density $w$.

\subsection{fANOVA of Piecewise-Constant Functions}

For $F$ which is piecewise-constant, we have a set of bins $\Omega_j$ for feature $j$. 
Let us assume that each $\Omega_j$ is finite and each bin is summarized by a single value, e.g. $\Omega_j = \{x_{j,1},\ldots,x_{j,n_{j}} \}$.
Then, the conditions \eqref{eq:integral_conditions} become:
\begin{subequations}
\begin{align}
    \forall~u \subseteq [d], &\forall~i~\in~u, \forall~X_{u\backslash i}, \quad
    \sum_{x_i \in \Omega_i}f_u(X_{u\backslash i}, X_i=x_i)\sum_{X_{-u}} w(X) = 0 \label{eq:sum_conditions}.
\end{align}
\end{subequations}
That is, if we represent each $f_u$ as a tensor of effect sizes, the fANOVA is recovered when every slice has mean zero with respect to a density $w$.

\section{Pure Interaction Effects}
\label{sec:pure}
We define \emph{pure interaction effect} as variance in the outcome which cannot be described by fewer variables:
\begin{definition}
Pure interaction effects of $X$ on $Y$ are:
\begin{subequations}
\begin{align*}
    &\{f_u(X_u)|u \subseteq [d] \} =  \nonumber \argmin_{\{g_u \in \Fc^u\}_{u\in [d]}} \int \Big(\sum_{u \subseteq [d]}g_u(X_u) - \E[Y|X] \Big)^2p(X)dX, \\
&\text{such that }\forall~u~\in~[d],
    \E[f_u(X_u)| X_v] = 0 \quad \forall~v~\subset u.
\end{align*}
\end{subequations}
\end{definition}
\noindent This is equivalent to the fANOVA decomposition of $\E[Y|X] = F(X)$ under $w(X) = p(X)$. 

\subsection{A Connection to Effect Coding}
\label{sec:alg:effect}
In the context of discrete features (e.g., Fig.~\ref{fig:xor}), fANOVA is equivalent to effect coding \citep{bech2005effects}. 
This unfolds the feature values into indicators, referred to as dummy variables or one-hot encoding. 
In linear regression with $X_1$ taking values in $\{0,1\}$ this translates to
\begin{align}
Y = \beta_0 + \beta_1 \mathcal{I}(X_1=0) + \beta_2 \mathcal{I}(X_1=1) + \epsilon.
\end{align}
For identifiability, we must drop a parameter. 
One common strategy is to remove the indicator for a ``reference'' value (e.g., setting $\beta_1 = 0$), in which case $\beta_0$ becomes the predictor for examples with the reference value. 
An alternative ``effect'' coding seeks to ensure that $\beta_0$ represents the average outcome by:
\begin{align}
Y &= \beta_0 + \beta_1 I(X_1=1) - \beta_1 I(X_1=0) + \epsilon.
\end{align}
For Boolean $X_1$, doing this translates to representing $X_1$ as a single column taking values in $\{-1,1\}$. 
For $X_1$ with values $\{v_1,\ldots,v_k\}$, the effects are given by $k-1$ columns with the $j$th having values $\mathcal{I}(X_1 = v_{j+1}) - \mathcal{I}(X_1 = v_1)$. 
When interactions are employed between discrete features, the interaction is then represented by the elementwise product of each pair of columns in the individual effects. 
For two Boolean features this exactly produces the XOR representation of Fig. \ref{fig:xor}d. 
More generally, the use of effect coding exactly corresponds to the fANOVA representation under a uniform weight function.

A natural question is how to extend effect coding to continuous features. 
In this paper, we propose to recover the fANOVA by purifying tree-based models. 
This exploits the power of tree-based models to partition continuous variables into discrete bins, providing a data-driven method of extending effect coding to continuous variables.

\section{Calculating fANOVA of Tree-Based Models}
\label{sec:algorithm}

For tree-based models, we have a tensor $T_u$ representing the effect sizes of each set of variables $u$. 
According to \eqref{eq:sum_conditions}, if these tensors can be ``purified" such that each 1-dimensional slice has mean zero, we recover exactly the fANOVA decomposition. 
Let 
\begin{equation}
    m(T_u, i, X_{u\backslash i}) = \sum_{x_i \in \Omega_i}f_u(X_{u\backslash i}, x_i)\sum_{X_{-u}} w(X) \label{eq:tensor_mean}
\end{equation}
be the weighted mean of the slice of $T_u$ representing effect sizes for different values of $X_i$ when $X_{u\backslash i} = x_{u\backslash i}$. 
For any $(u, i)$, we also have a corresponding $T_{u\backslash i}$ (letting $T_{\emptyset}$ be the tensor representing the overall model intercept). 
Because the model predictions are generated by summing the effects across all $T$, we can move any value from $T_u$ into $T_{u\backslash i}$ without changing the model predictions. 
In particular, we can move $m(T_u, i, x_{u\backslash i})$ into $T_{u\backslash i}$ to generate a 1-dimensional slice of $T_u$ with mean zero without adjusting the output of the overall model. 
We refer to this as ``mass-moving" between $T_u$ and $T_{u\backslash i}$.

\begin{algorithm}[ht]
    \begin{algorithmic}[1]
\Require $T, w, u, \Omega$ 
\Comment{Will purify $T_u$ so that every slice has zero-mean according to weighting $w$}
\State end $\gets$ False
\While {pure $\neq$ True}
    \State pure $\gets$ True
    \For {$i \in u$}
        \For {$x_{u\backslash i} \in \Omega_{u\backslash i}$}
            \State $m^0 \gets m(T_u, i, x_{u \backslash i})$
            \Comment Calculate according to Eq.~\eqref{eq:tensor_mean}
            \If {$m^0 \neq 0$}
                \Comment There is still mass to move.
                \State pure $\gets$ False
                \State $T_{u}[x_{u \backslash i}, :] \gets T_{u}[x_{u \backslash i}, :] - m^0$
                \State $T_{u \backslash i}[x_{u \backslash i}] \gets T_{u \backslash i}[x_{u \backslash i}] + m^0$
            \EndIf
        \EndFor
    \EndFor
\EndWhile \\
\Return $T$ 
\end{algorithmic}
\caption{Purify-Matrix \label{alg:helper}}
\end{algorithm}

\begin{algorithm}[ht]
    \begin{algorithmic}[1]
    \Require $T, w, \Omega$
    \Comment{$T$ is the set of tensors, $\Omega$ is the values, $w$ is the weighting}
    \State $order \gets $sort\_descending$([|T|])$ \Comment{Arrange all potential sets $u$ in order of decreasing size}
    \For {$u \in order$}
        \State $T \gets $Purify-Matrix$(T, w, u, \Omega)$
    \EndFor
    \State 
    \Return {$T$}
    \end{algorithmic}
\caption{Purify \label{alg:purify}}
\end{algorithm}

This suggests an algorithm for calculating the fANOVA for tree-based models. 
By iteratively removing the means of slices of $T_u$, we can generate a $T_u$ which satisfies \eqref{eq:sum_conditions} without changing the model's outputs. 
We can iteratively purify all $T_u \in T$ by this procedure, cascading effects from high-order interactions into low-order interactions, from low-order interactions into main effects, and finally from main effects into the global intercept. 
We call this algorithm ``mass-moving" (Alg.~\ref{alg:purify}) because it iteratively moves mass from higher-order interactions to lower-order interactions until no mass remains to be moved. 
At convergence, it exactly recovers the fANOVA decomposition.
This algorithm can be implemented in under 100 lines of Python. An implementation is available at \url{https://github.com/microsoft/interpret}.

In contrast to other algorithms for calculating the fANOVA which rely on optimization of orthogonal basis functions \citep[e.g.][]{hooker2007generalized,chastaing2012}, our algorithm uses the tree structure to recover the exact decomposition. 
This avoids challenges of optimizing functions on correlated variables. 
While this paper is focused on the \emph{implications} of the decomposition, and thus we study the results of decomposing tree-based models, in principle we could apply this algorithm to any $F$ by first estimating a piecewise-constant $\hat{F}$ (e.g. with an adaptive tree-based model).

\subsection{Convergence and Correctness}
\label{sec:algorithm:convergence}
By the uniqueness of the fANOVA, Alg.~\ref{alg:purify} is correct if and only if it converges to produce tensors with zero-mean slices. 
Since Alg.~\ref{alg:purify} operates on the tensors in order of decreasing dimension, it suffices to check that Alg.~\ref{alg:helper} converges to produce a tensor with zero-mean slices for any input. 
To see that this is indeed the case, let us examine the means of slices over a run of Alg.~\ref{alg:helper} for a matrix $T_{a,b}$ representing the effect of the interaction of two variables $X_a \in \Omega_a$ and $X_b \in \Omega_b$. 
For simplicity, we are considering only a matrix representing an interaction between two variables; this proof extends to tensors as well since the fANOVA is defined over one-dimensional slices. 
In the following, we use the shorthand notation $w_{i,j} = w(i, j)$ (for $i \in \Omega_a$ and $j \in \Omega_b$), and assume that $w$ has been normalized so that $\sum_{i \in \Omega_a}\sum_{j \in \Omega_b}w_{i,j} = 1$.

Let $t$ be an iteration counter which alternates between zeroing row and columns (i.e., the number of times line 4 of Alg.~\ref{alg:helper} has been passed). 
At each $t$ we have a matrix $T_{a,b}^t$ and can define: 
\begin{subequations}
\begin{align}
    c_j^t &= \sum_{i \in \Omega_a} w_{i,j}T_{a,b}^t[i,j] \\
    r_i^t &= \sum_{j \in \Omega_b} w_{i,j}T_{a,b}^t[i,j] \\
    M^t &= \sum_{i \in \Omega_a}\sum_{j \in \Omega_b} w_{i, j}(\abs{r_i^t} + \abs{c_j^t})
\end{align}
\end{subequations}
That is, $M^t$ is the unpurified mass at iteration $t$; when $M^t = 0$ the algorithm has converged to a matrix with zero-mean at every slice with non-zero probability. 
Then, for any $w$ which has equal weighting along the row or column dimensions, the algorithm converges in a single iteration:
\begin{theorem}
For any $T_{a,b}, \Omega$, if $w_{i,j}=w_{i,j'}~\forall~i,j'$: \label{thm:uniform}
\begin{align}
    M^{t} = 0 \quad \forall~t~ \geq 2
\end{align}
\end{theorem}
\noindent
This means the algorithm converges in a single pass for many simple distributions, a fact familiar to data scientists who often ``double-center" design matrices with uniform weighting over samples and features. 
We also have rapid convergence of Alg.~\ref{alg:helper} for generic non-degenerate $w$:
\begin{theorem}
For any $T_{a,b}, w, \Omega$, for any $\epsilon > 0$ \label{thm:convergence}
\begin{align}
    M^{t} \leq \epsilon \quad \forall~t~ \geq \tau(\epsilon)
\end{align}
where $\tau(\epsilon) = \log_2\big( \frac{M^0}{\epsilon} \big)$. 
\end{theorem}
\noindent 
That is, Alg.~\ref{alg:purify} converges to the fANOVA decomposition with tolerance $\epsilon$ in $\mathcal{O}\big(\log(M^0) - \log(\epsilon))$ iterations for each interaction tensor. 
This theorem is proved in Sec~\ref{sec:appendix:proof} of the Supplement. 
Empirically, we observe that most of the mass is moved in very few iterations (Section~\ref{sec:alg:measuring} and Section~\ref{sec:appendix:convergence} of the Supplement).

\subsubsection{fANOVA Properties}
Since we have shown that Alg.~\ref{alg:purify} converges to the fANOVA decomposition, the uniqueness of the decomposition provides several useful corollaries. 
First, permutation of the rows and columns does not change the purified interaction effect:
\begin{corollary}
For any permutation $P$ with inverse $P'$, 
\begin{align}
    \text{Purify-Matrix}(\{T_a, T_b, T_{a,b}\}, w, \{a, b\}, \Omega) = P'(\text{Purify-Matrix}(P(T_a, T_b, T_{a,b}), w, \{a, b\}, \Omega)).
\end{align}
\end{corollary}
\noindent This gives two convenient conditions: (1) re-encoding the order of nominal variables does not change the interaction effects, and (2) Alg.~\ref{alg:helper} can iterate over the slices of a tensor in any order.

Second, interaction purification is a linear operator:
\begin{corollary}
For any interaction matrix $T_{a,b} = \alpha_1A_1 + \ldots + \alpha_nA_n$, where $\sum_{i=1}^n\alpha_i = 1$:
\begin{align}
    \text{Purify-Matrix}&(\{T_a, T_b, T_{a,b}\}, w, \{a, b\}, \Omega) \nonumber \\
    &= \alpha_1\text{Purify-Matrix}(\{T_a, T_b, A_1\}, w, \{a, b\}, \Omega) \nonumber \\
    & + \ldots + \alpha_n\text{Purify-Matrix}(\{T_a, T_b, A_n\}, w, \{a, b\}, \Omega).
\end{align}
\end{corollary}
\noindent
This means that purification can be run equivalently before or after bootstrap aggregation.

\subsection{Estimating \texorpdfstring{$w$}{w}}
\label{sec:estimating}
Defining interaction effects via weighted fANOVA makes it clear that effects can only be understood in conjunction with a data distribution. 
The correct $w(x)$ under which to understand effects is the true data distribution $p(x)$; however a fundamental challenge of machine learning is to estimate $p(x)$ from limited data. 
In this paper, we use three simple estimators of piecewise-constant densities:
\begin{itemize}
    \item \bf{Uniform:} $\hat{w}_{\text{unif}}(x_{-u}) \propto 1$
    \item \bf{Empirical:} $\hat{w}_{\text{emp}}(x_{-u}) \propto \sum_{x^i \in X_{\text{train}}}\Ic_{\{x_{-u}^i = x_{-u}\}}$
    \item \bf{Laplace:} $\hat{w}_{\text{lap}}(x_{-u}) \propto \hat{w}_{\text{unif}}(x_{-u}) + \hat{w}_{\text{emp}}(x_{-u})$ 
\end{itemize}
As we see in the experiments, the choice of distribution can change (sometimes dramatically) the purified effects. 
Thus, selection of $\hat{w}(x)$ is a critical step in model interpretation and we look forward to future work which improves estimation of $p(x)$.

\subsection{Measuring Convergence of the Purification Algorithm}
\label{sec:alg:measuring}
\begin{figure}[ht]
    \centering
    \begin{subfigure}[t]{0.31\columnwidth}
        \vskip 0pt
        \centering
        \includegraphics[width=\textwidth]{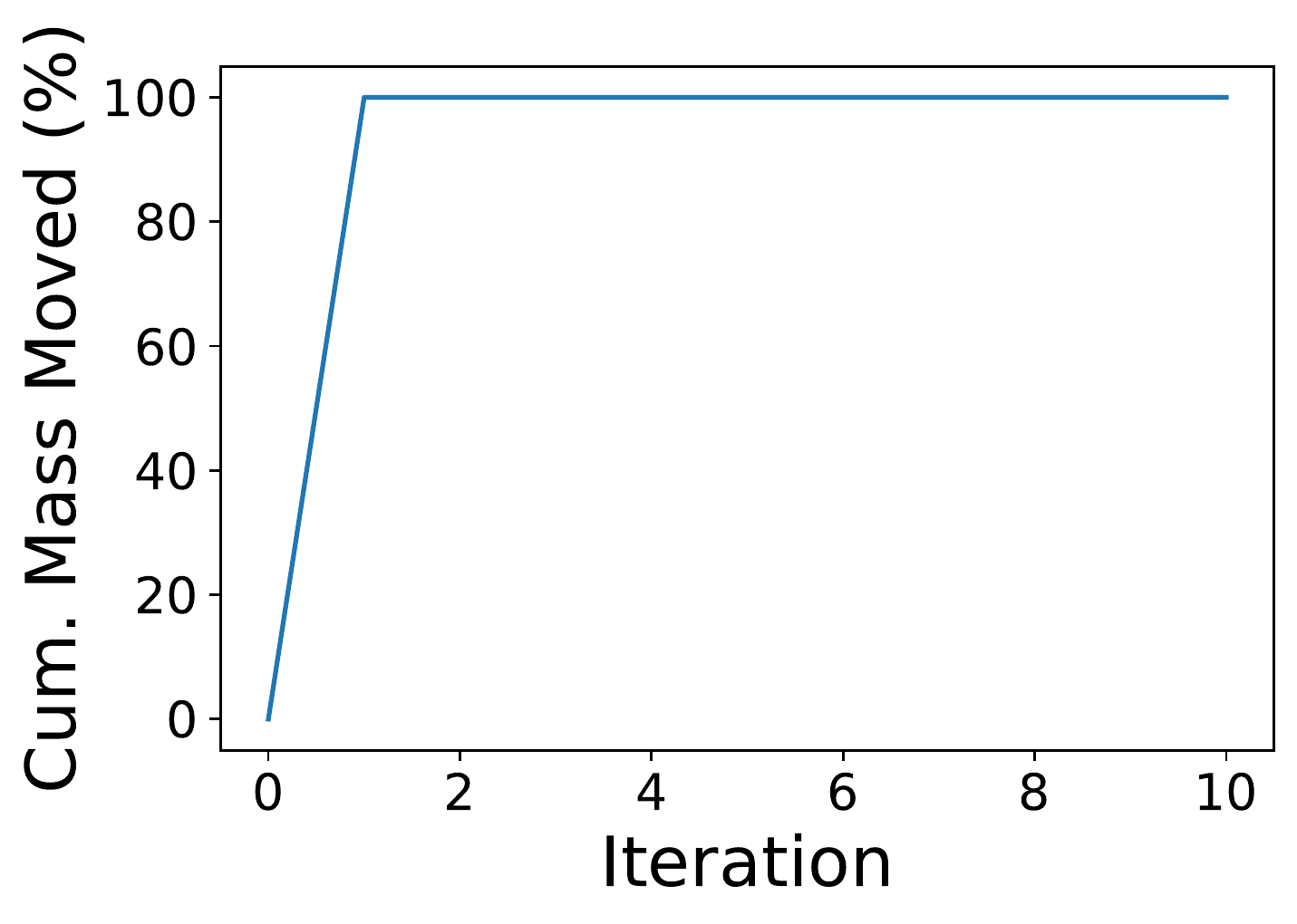}
        \caption{Uniform density, $\sigma=1$, $P=100$}
    \end{subfigure}
    ~
    \begin{subfigure}[t]{0.31\columnwidth}
        \vskip 0pt
        \centering
        \includegraphics[width=\textwidth]{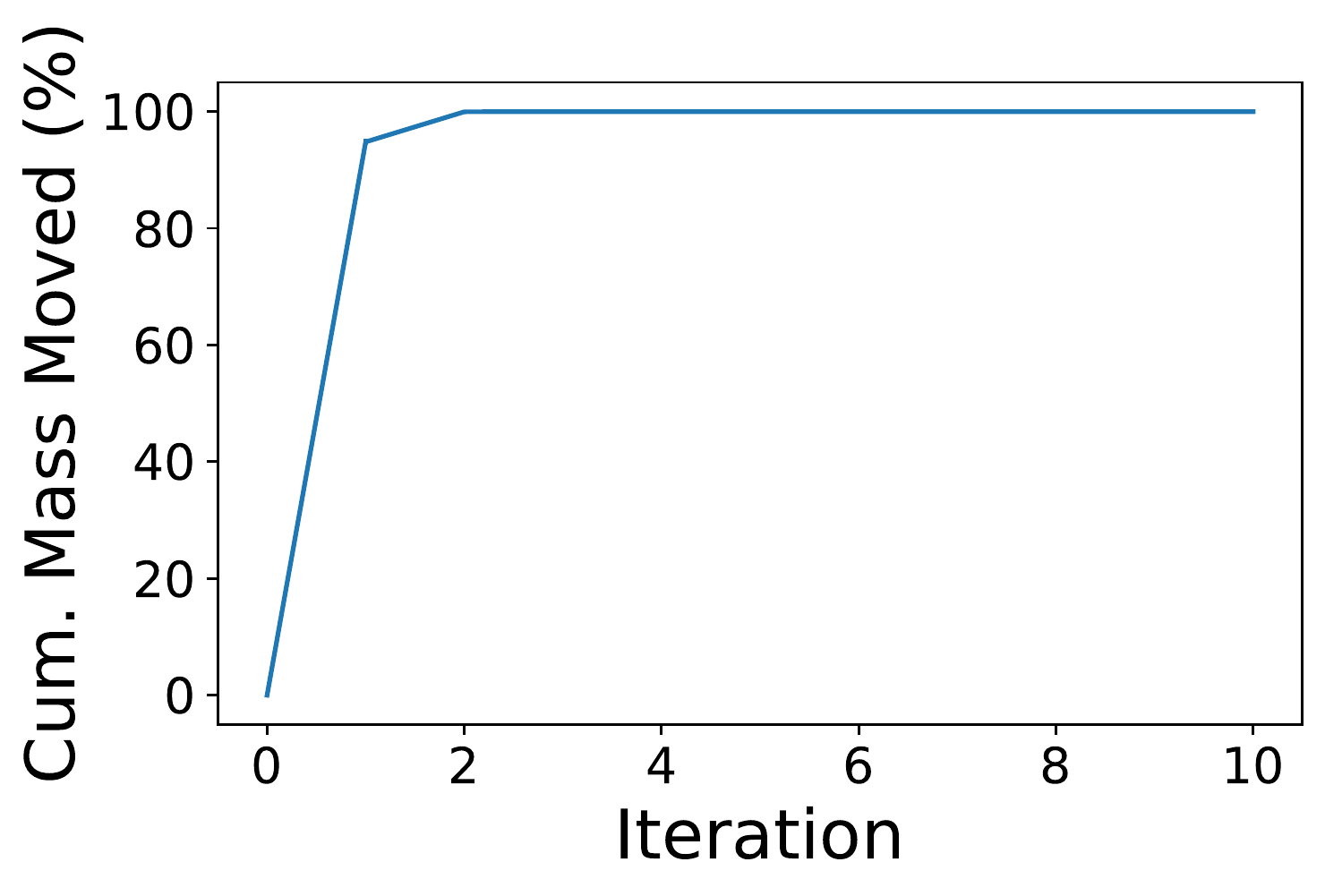}
        \caption{Density drawn from multivariate normal $\sigma=10$, $P=100$}
    \end{subfigure}
    \caption{Convergence of Algorithm~\ref{alg:helper}. Errorbars indicate (mean $\pm$ std) over 100 experiments of drawing tensor and weight values. The mass-moving algorithm converges in a very small number of iterations for both the uniform and the random weighting. \label{fig:convergence_representative}}
\end{figure}

Purification by the mass-moving algorithm converges in a very small number of iterations. 
We examine this behavior empirically by generating tensors $T \sim N(0, \sigma I)$ with weight values either: (1) uniform distribution: $w \propto 1$ or (2) drawn from a multivariate normal distribution: $w \sim N(0, \sigma I)$ of dimension $P$. 
Representative results are shown in Figure~\ref{fig:convergence_representative}; results for a variety of settings of $\sigma$ and $P$ are shown in Figures~\ref{fig:convergence_uniform} 
and ~\ref{fig:convergence_random} in the Supplement. 
In all cases, we see that the mass-moving algorithm moves almost all of the mass in the first iteration. 
In the case of the uniform weight distribution, we confirm that the mass-moving algorithm takes only a single iteration (per row/column) to convergence. 
These results show that the mass-moving algorithm can scale to purify large models.

\section{Experiments}
\label{sec:experiments}
We examine the implications of purification on models trained from several datasets. 
To do so, we use two types of additive models with interactions. 
Both of these models are tree-based ensembles, from which we recover the set of effect tensors $T$ by summing the effect tensors of each tree in the forest. 

The first model is a constrained form of Extreme Gradient Boosted (XGB) Forests \citep{chen2016xgboost}, an extremely popular model for tabular data. 
By limiting the depth of each tree to a single split (boosted stumps, referred hereafter as XGB), this is a GAM without interactions. 
If we allow the trees to have depth 2, this model (XGB2) is a GAM with pairwise interactions. 
The XGB2 model was not designed to prefer main effects over interactions. 
As we see in the experiments, this means that purification induces large changes in the interpretation of XGB/XGB2 models.

The second model we use is the GA2M model \citep{lou2013accurate} implemented in \citealt{nori2019interpretml}.
The GA2M algorithm allows users to specify the number of interactions to estimate; when this value is set to 0 we refer to this algorithm as GAM. 
GA2M was designed with a two-stage estimation procedure to fit main effects before fitting interactions in order to make mains as strong as possible and prevent main effect from leaking into interactions. 
This two-stage training procedure reduces the mass that needs to be moved by purification; nevertheless on average the purification process also improves the main effects learned by GA2M models.

Our results show that model interpretations change significantly from purification and that the choice of data distribution used for purification is important to understand model implications.

\begin{figure}[tb]
    \centering
    \begin{subfigure}[b]{0.45\columnwidth}
        \centering
        \includegraphics[width=0.9\textwidth]{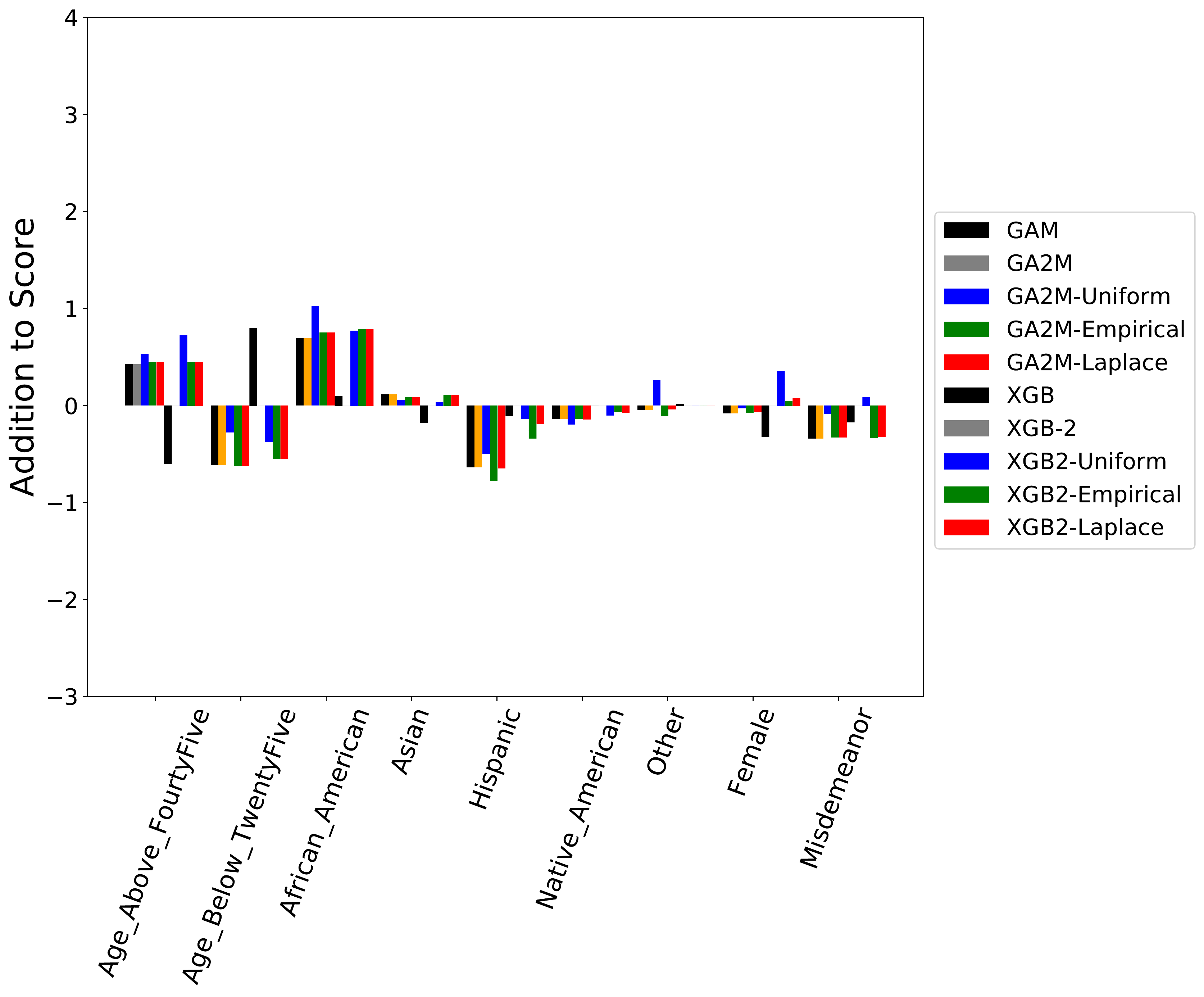}
        \caption{Prediction of Recidivism}
    \end{subfigure}
    ~
    \begin{subfigure}[b]{0.45\columnwidth}
        \centering
    \includegraphics[width=0.9\columnwidth]{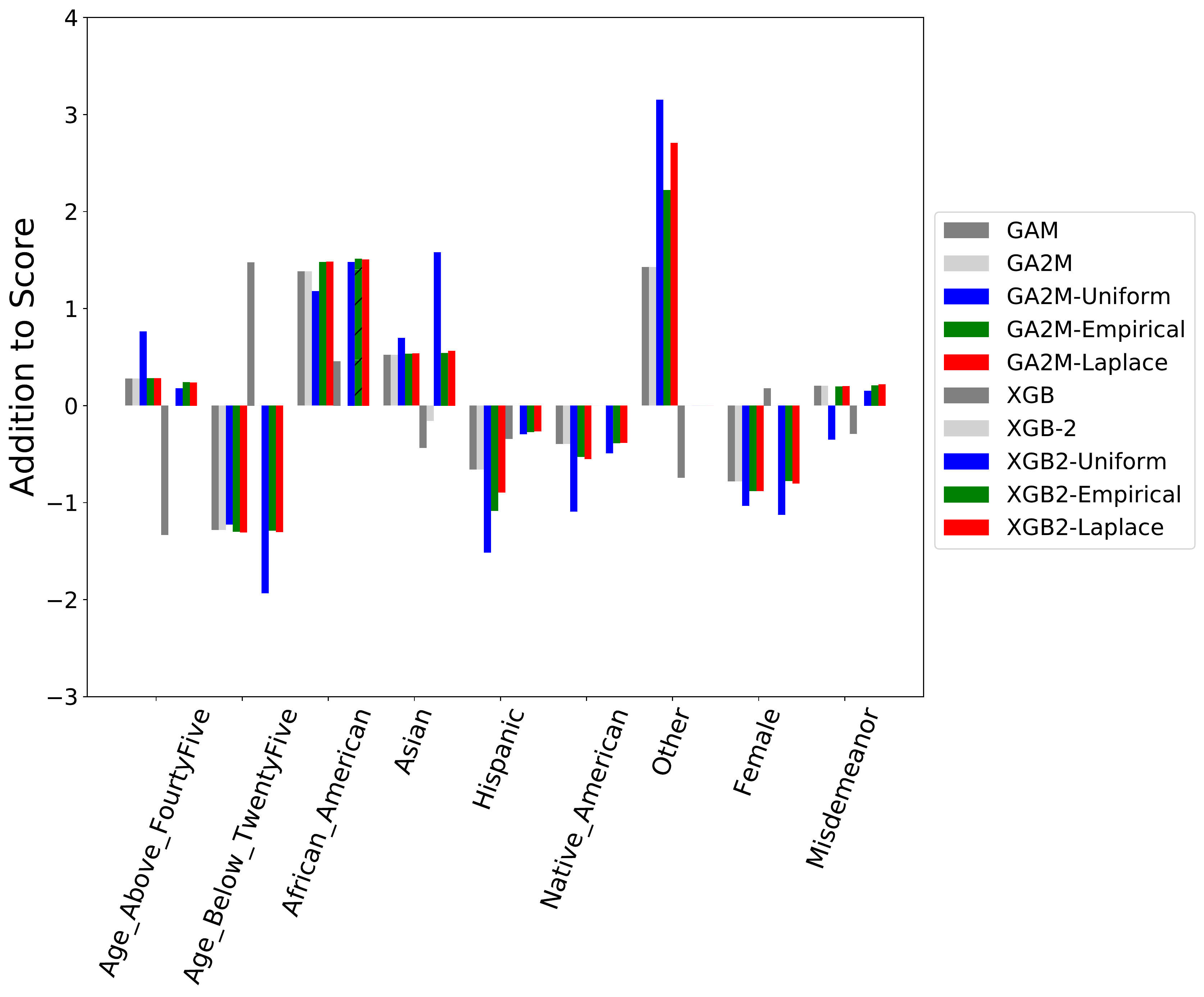}
        \caption{Prediction of COMPAS Score}
    \end{subfigure}
    \caption{Main effects of additive models with interactions trained to predict the (a) ground-truth recidivism and (b) COMPAS risk score. The implications of the main effects depend on the model class, the use of purification, and the distribution used for purification.}
    \label{fig:compas}
\end{figure}

\subsection{COMPAS}
\label{sec:exp:compas}
The Correctional Offender Management Profiling for Alternative Sanctions (COMPAS) system is a model for predicting recidivism risk that is used to guide bail decisions. 
The high stakes of this system make it crucial to ensure that the algorithm treats individuals fairly -- understanding how COMPAS makes predictions is of societal importance.

In 2016, the investigative journalism firm Propublica organized and released recidivism data on defendants in Broward County, Florida along with the correponsding predictions from the COMPAS model\footnote{\url{https://github.com/propublica/compas-analysis/}}. 
Analyses of this dataset have sparked controversy, with different investigations coming to different conclusions about algorithmic bias \citep{dieterich2016compas,feller2016computer,tan2017detecting}. 
Here, we ask whether the conclusions regarding algorithmic bias are changed by purification, and how much the choice of sample distribution changes the interpretation of the model.

To answer this question, we train an additive model with interactions to mimic the COMPAS model, as in \cite{tan2017detecting}. 
As shown in Fig.~\ref{fig:compas}, the interpretations of main effects in the COMPAS dataset are changed significantly by the purification process. 
The magnitude -- and occasionally the sign -- of the effects are changed by the selection of data distribution. 
In particular, the sign of many mains learned by XGB (gray bars) are opposite the signs for those mains learned by GAMs, or GA2Ms and XGB2 after purification: the use of purification with XGB2 yields mains that are much more consistent with what other models learn compared to mains learned directly by XGB. 
Also, there is significant variation in the strength of the mains (though not the signs) depending on the data distribution used for the purification process. 
Both the learning algorithm and purification distribution are important for meaningful audits of the COMPAS model.

\begin{figure*}[tp]
    \centering
    \begin{subfigure}[t]{\textwidth}
        \centering
        \begin{subfigure}[t]{0.55\columnwidth}
            \vskip 0pt
            \centering
            \includegraphics[width=\textwidth]{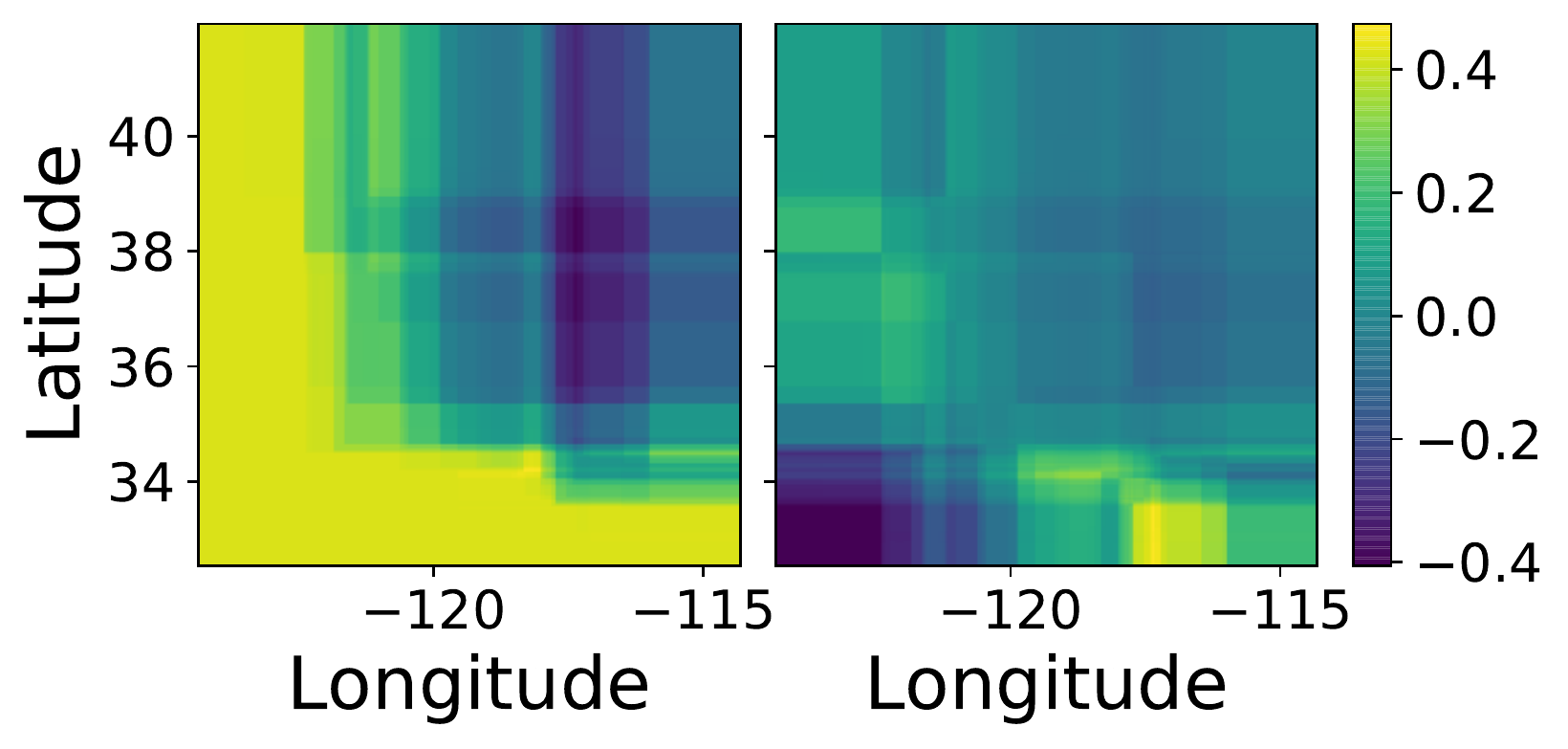}
        \end{subfigure}
        ~
        \begin{subfigure}[t]{0.25\columnwidth}
            \vskip 0pt
            \centering
            \includegraphics[width=\textwidth]{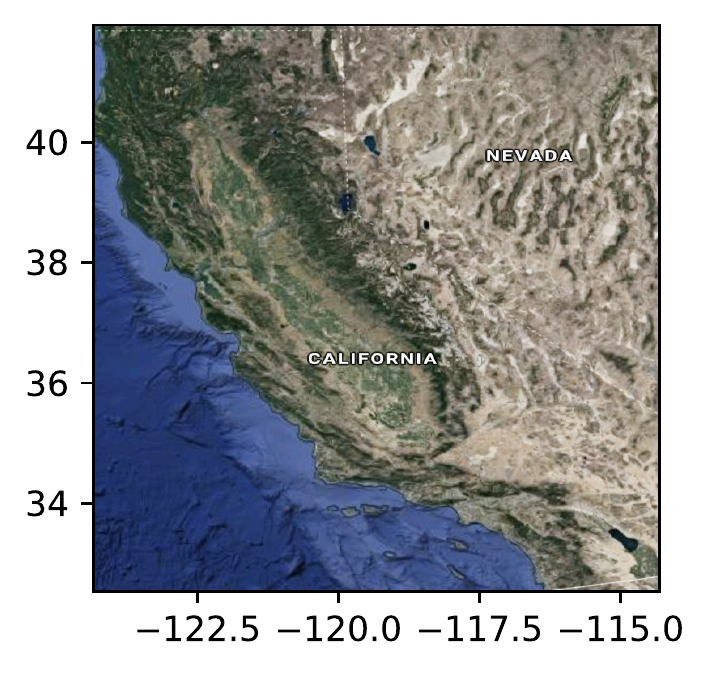}
        \end{subfigure}
    \end{subfigure}
    \caption{Interaction of the Latitude/Longitude features in an XGB2 model trained on the California housing data. 
    The left pane is the unpurified interaction, the middle pane is the purified interaction, and the right is the map of California from which samples were drawn. 
    Purification sorts out the influence from the Los Angeles and the San Francisco metro areas. }
    \label{fig:cali_lat_long}
\end{figure*}

\subsection{California Housing}
\label{sec:exp:cali}
A canonical machine learning dataset, and the task used in the original development of weighted fANOVA \citep{hooker2007generalized}, is the California Housing dataset \citep{pace1997sparse}. 
This dataset was derived from the 1990 U.S. census to understand the influence of community characteristics on housing prices. 
The task is regression to predict the median price of houses in each district in California.

In Fig.~\ref{fig:cali_lat_long}, we see the interaction of latitude and longitude on housing prices. 
The unpurified effects indicate that the most expensive real estate lies in the Pacific Ocean; after purification this problem goes away and we see that it arose from the influence of the Los Angeles and San Francisco metropolitan areas. 
This result is similar to the fANOVA decomposition in \cite{hooker2007generalized} (see Fig.5 within); however, our approach is able to recover these pure interaction effects from any model, rather than constrained to orthogonal basis functions. This enables our approach to use methods which adaptively split variables (such as the gradient-boosted trees of XGB2), leading to more refined density estimation than the grid used in \cite{hooker2007generalized}.

\begin{figure}[tb]
    \centering
    \begin{subfigure}{0.5\columnwidth}
        \centering
        \includegraphics[width=0.48\textwidth]{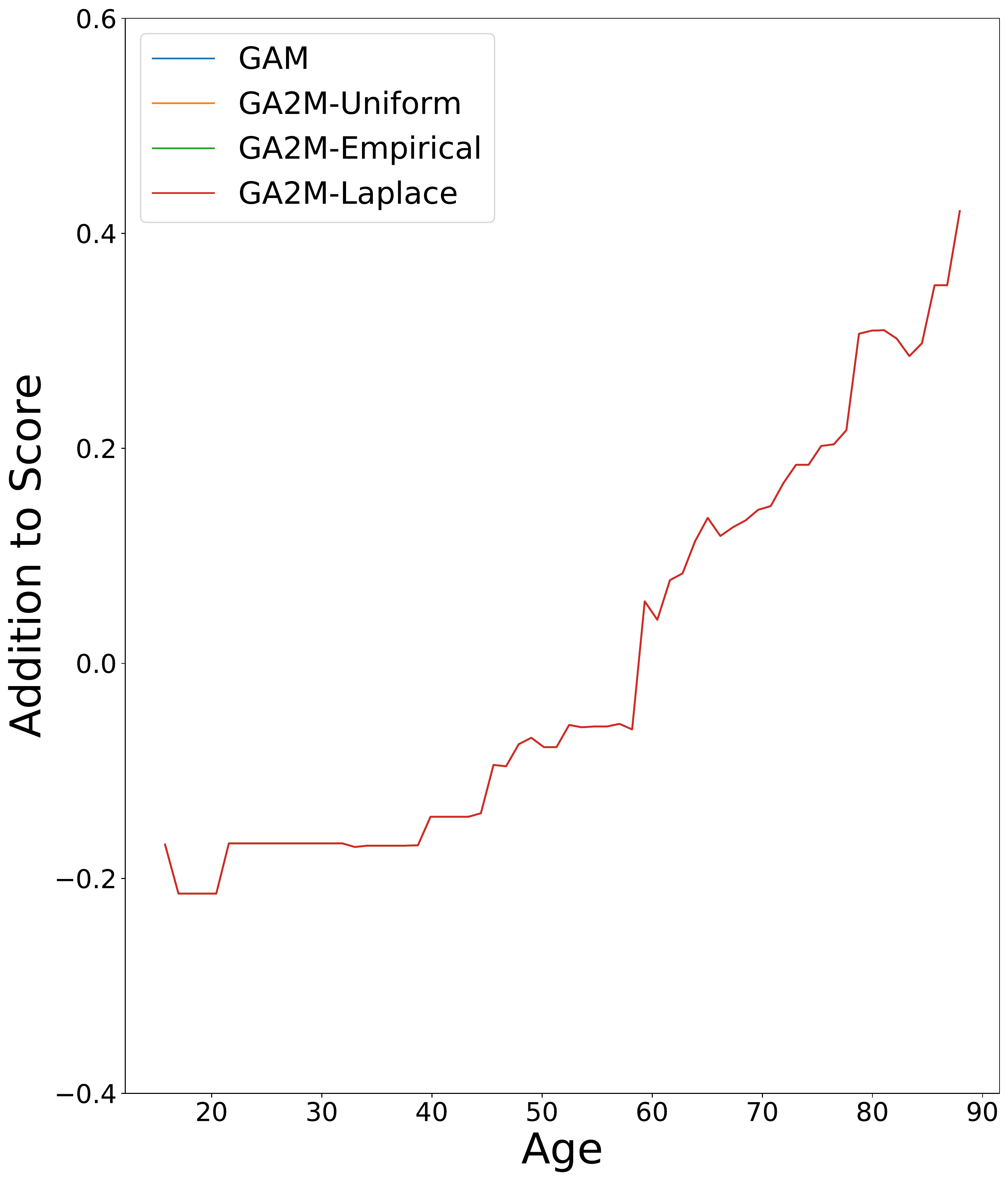}
        ~
        \includegraphics[width=0.48\textwidth]{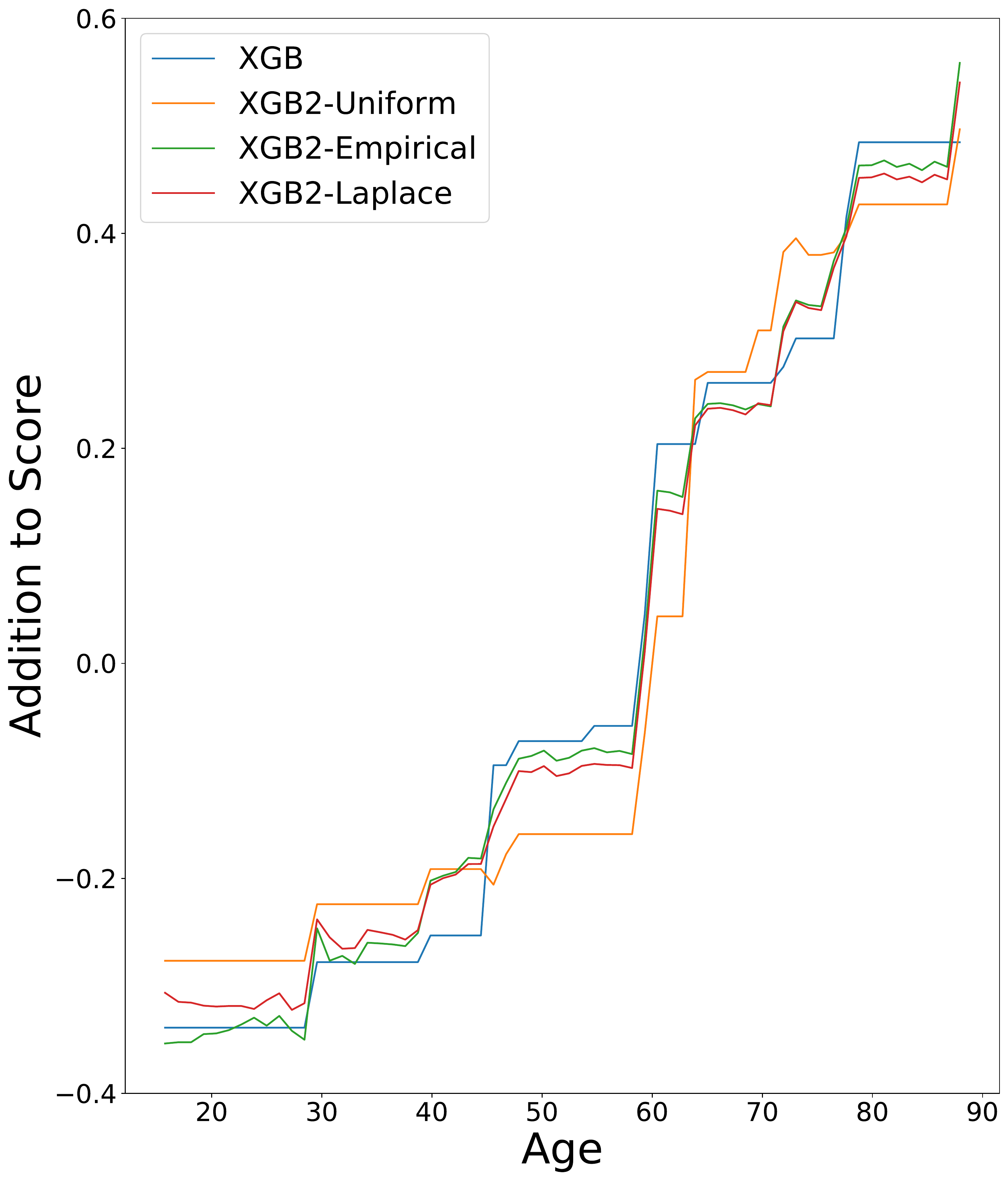}
        \vspace{-5pt}
        \caption{Age}
    \end{subfigure}
    ~
    \begin{subfigure}{0.5\columnwidth}
    \vskip 5pt
        \centering
        \includegraphics[width=0.48\textwidth]{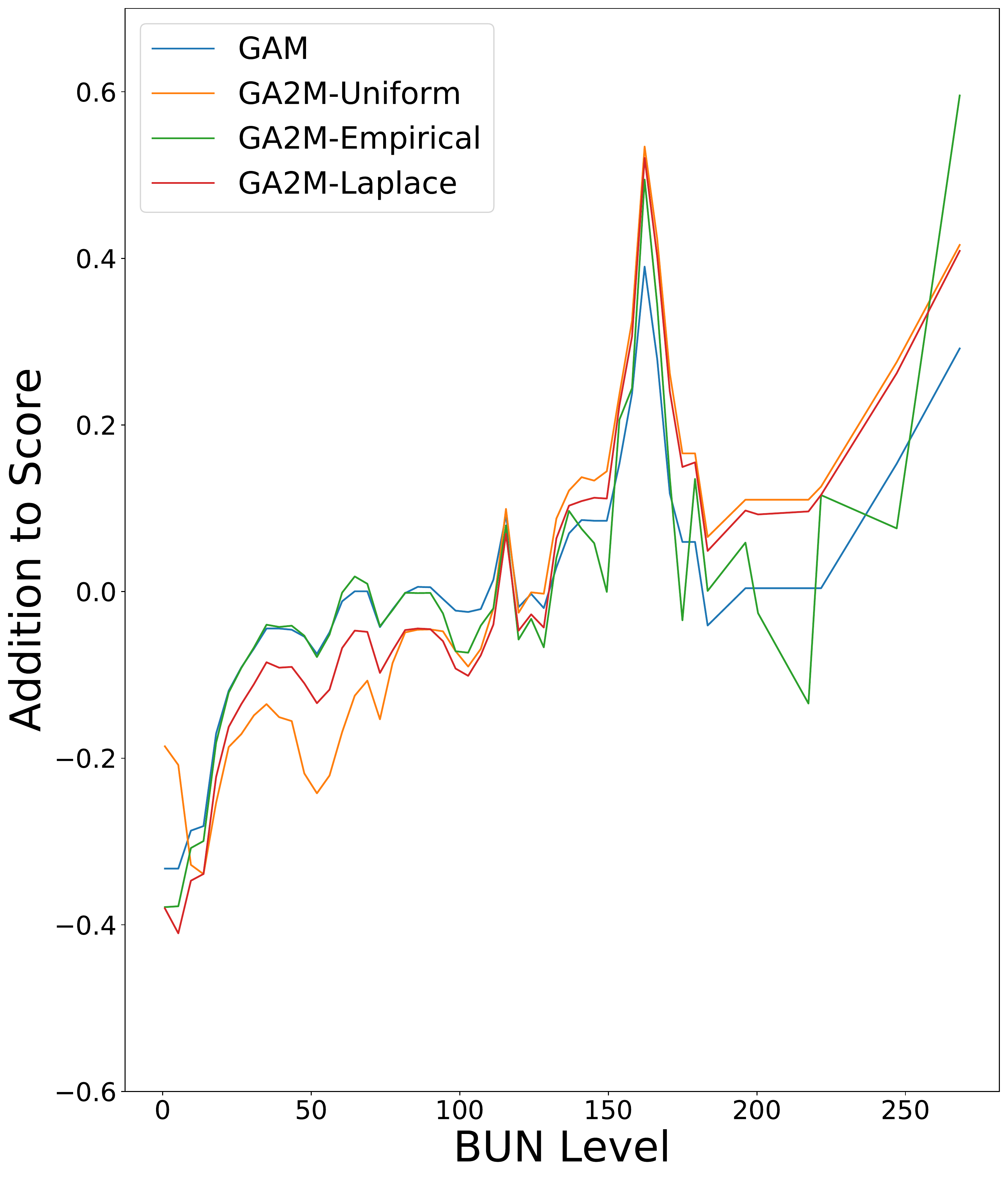}
        ~
        \includegraphics[width=0.48\textwidth]{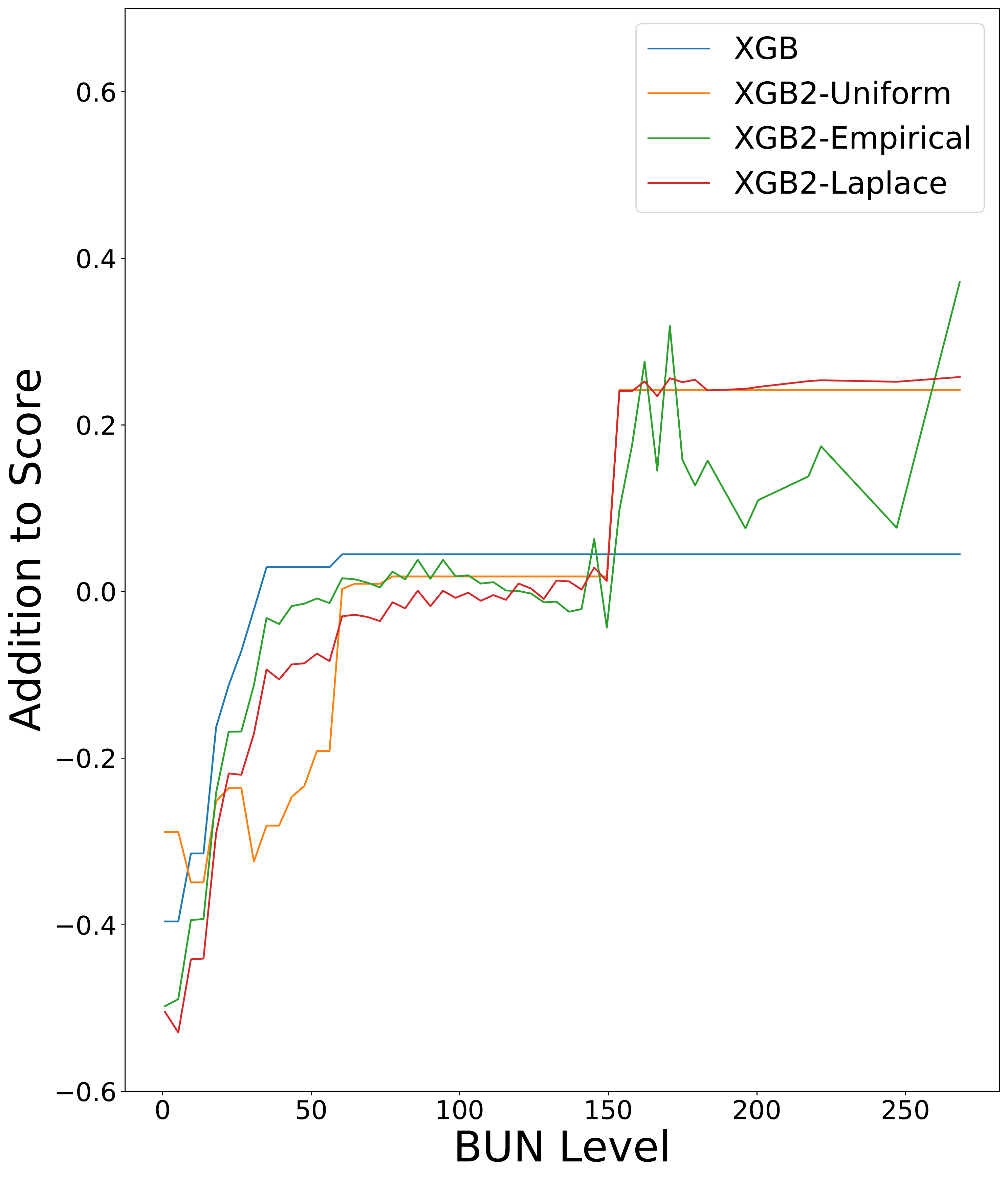}
        \caption{Blood Urea Nitrogen (BUN) level}
    \end{subfigure}
    \caption{Main effects of models trained to predict mortality in MIMIC-III before and after purification.}
    \label{fig:mimic}
\end{figure}

\subsection{MIMIC-III}
\label{sec:exp:mimic}
MIMIC-III \citep{johnson2016mimic} is a medical dataset of lab tests and outcomes for patients in the intensive care unit (ICU). 
The classification task is to predict mortality in the current hospital stay. 
In this experiment, we investigate the reliability of risk curves by examining their consistency after purification with different sample distributions.

A representative sample of main effect curves are shown in Fig.~\ref{fig:mimic}. The upper left graph (the main effect of Age for GAM/GA2Ms) shows imperceptible change due to purification. 
As a result, the selection of distribution used for purification does not make a large difference in this case. 
For the XGB/XGB2 models, however, this story is more complex (upper right graph). 
The XGB2 model is not designed to prioritize main effects over interaction effects, so purification makes non-negligible impact and the selection of a distribution can change model interpretation.

These differences are magnified for the variable blood urea nitrogen (BUN) which participates in a large number of interactions. 
Even though the GA2M algorithm was designed to estimate interactions based only on residuals after estimating the main effects, it is apparent that the interaction terms still capture some main effects because mass-moving significantly alters the main effect of this variable. 
As a result, the purified risk curves can produce different interpretations for different distributions (e.g., the curve of the risk graph at BUN near 50).

Purification does not change the overall model, so any excessive granularity in the purified main effects must have been hiding in the interaction effects. 
This leads us to believe that tree-based models tend to estimate high-variance interaction effects, and that regularizing these interaction effects could improve predictive accuracy. 

\section{Discussion}
\label{sec:discussion}

\subsection{The Mystery of \texorpdfstring{$\log(X_1X_2)$}{log(X1X2)}}
\label{sec:discussion:log}

In this section we use the notion of pure interaction effects to revisit a classic interaction puzzle: the $\log$ function.  
A classic way of representing an interaction between $X_1$ and $X_2$ is $Y=X_1X_2$. 
If we sample data for $X_1$ and $X_2$ uniformly on the the interval $(0,1]$, the mass-moving purification algorithm shows that $Y=X_1X_2$ has the following XOR-like interaction and linear main effects:

\begin{figure}[H]
    \centering
    \includegraphics[width=0.2\columnwidth]{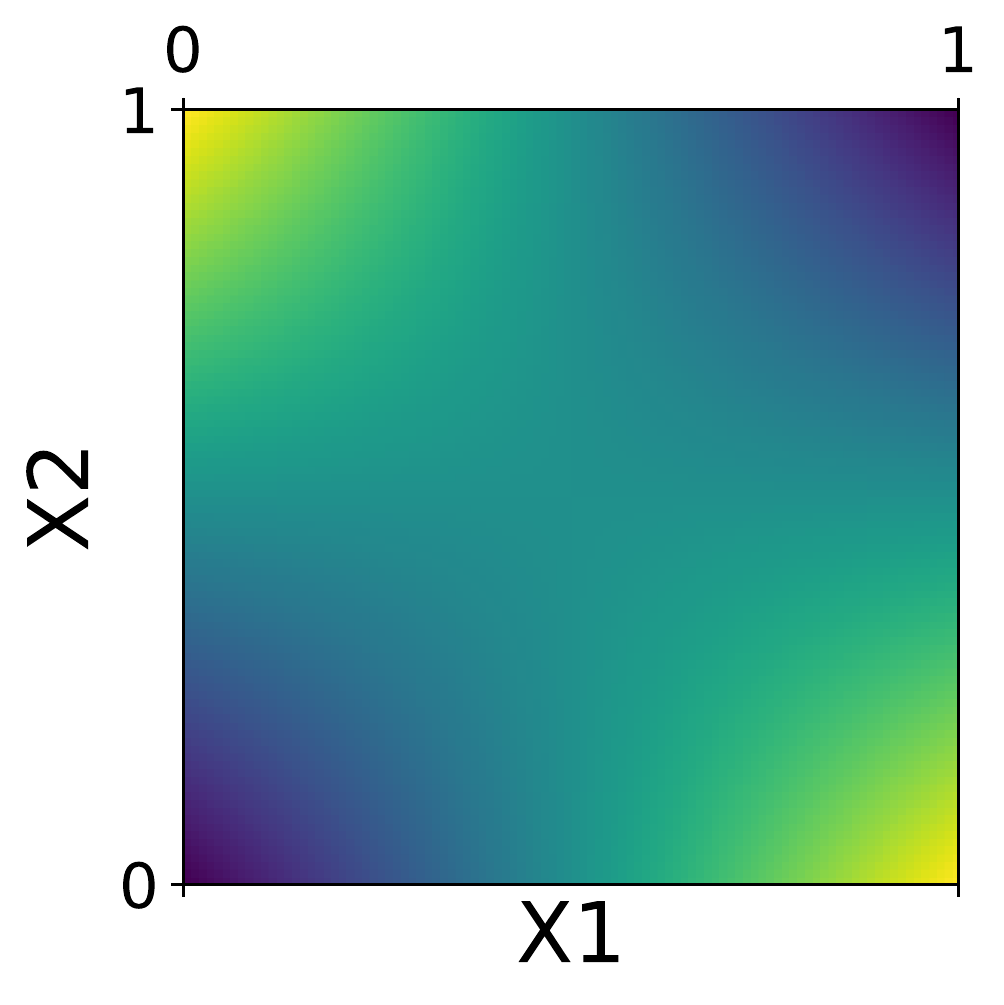}
    ~
    \includegraphics[width=0.18\columnwidth]{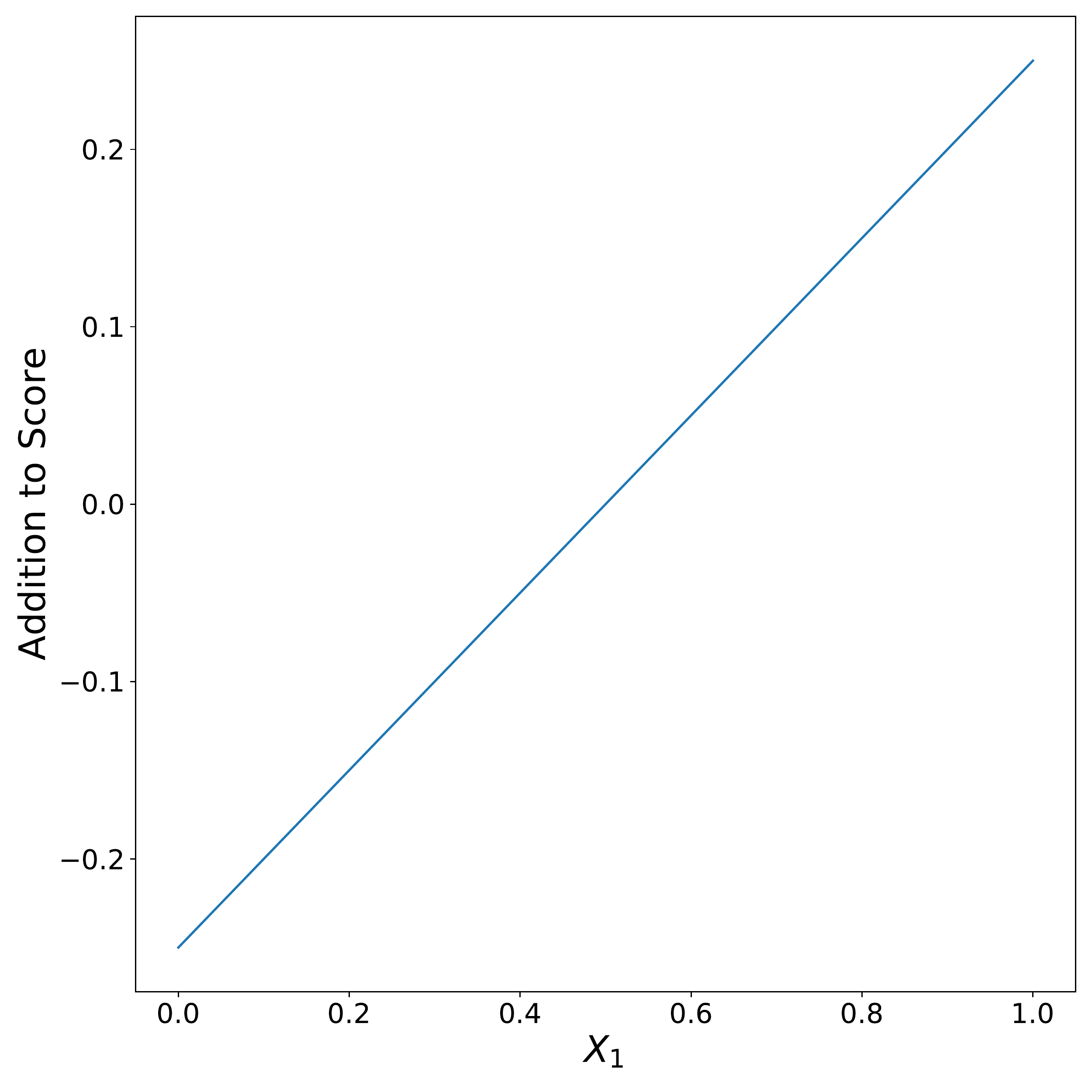}
    ~
    \includegraphics[width=0.18\columnwidth]{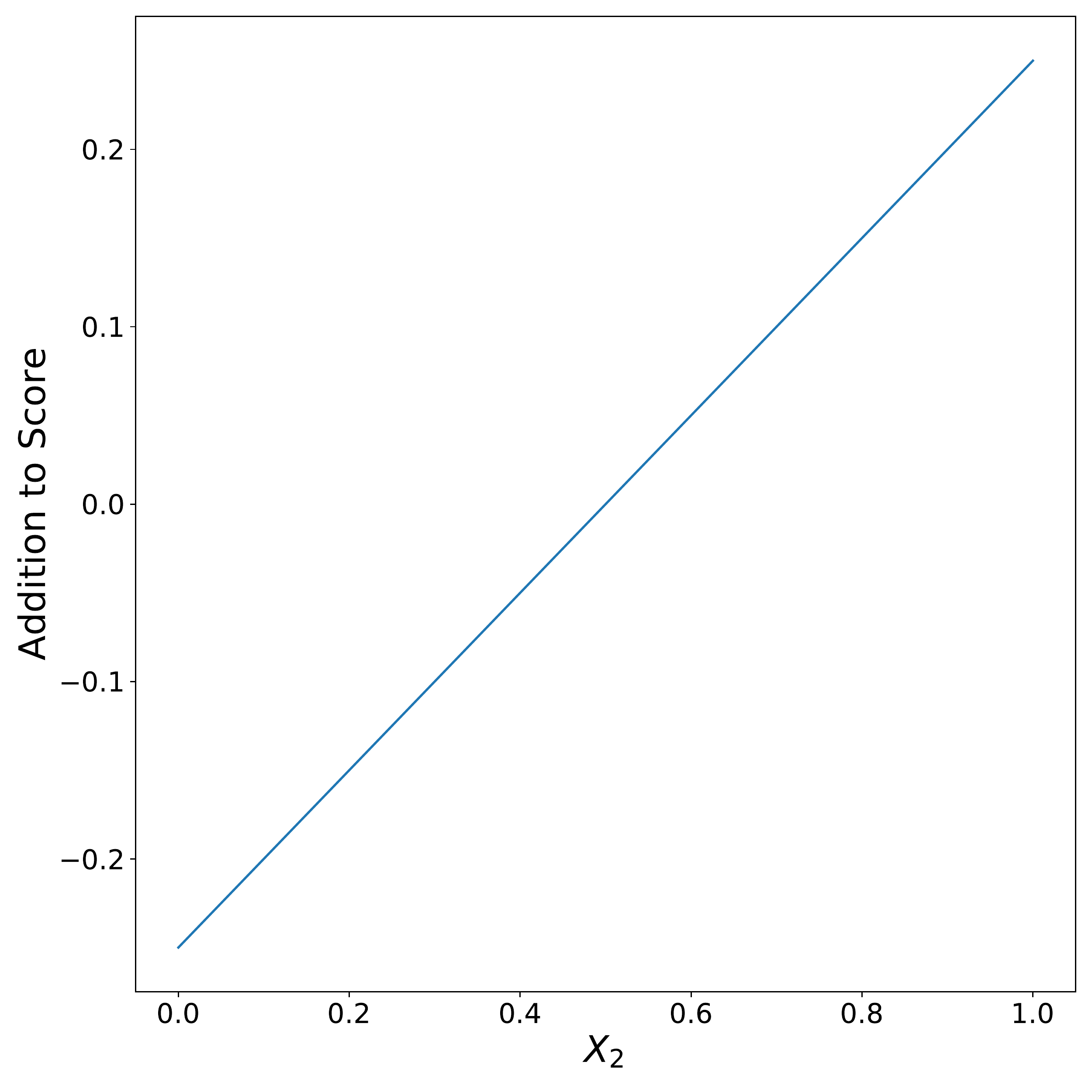}
    \caption{The pure interaction and main effect of $Y=X_1X_2$.}
    \label{fig:mult_purified}
\end{figure}

However, if we model the logarithm of $Y$, we get  $\log(Y)=\log(X_1X_2)=\log(X_1)+\log(X_2)$. 
That is, applying $\log(\cdot)$ to the interaction $X_1X_2$ appears to break the interaction and yield a model that is purely additive in the $\log(X_1)$ and $\log(X_2)$. 
It is surprising that applying a simple monotone function to the product $X_1X_2$ can make the interaction between $X_1$ and $X_2$ disappear. 

Does purification by mass-moving account for this? 
Yes -- according to our definition of pure interaction effects as satisfying the fANOVA, $\log(X_1X_2)$ is not an interaction effect at all. 
To test that our mass-moving procedure recovers the correct decomposition, we generate data according to the model:
\begin{equation}
    Y = (1-\lambda)\log(X_1X_2) + \lambda(X_1X_2),
    \label{eq:log_model}
\end{equation}
which allows us to control how much the function applied to $X_1X_2$ behaves like $\log()$ for $\lambda\approx 0$ vs. multiplicative interaction for $\lambda \approx 1$. 
By varying $\lambda \in [0, 1]$, we can examine the ability of Alg.~\ref{alg:purify} to distinguish these effects.

\begin{figure}[H]
    \centering
    \vskip 0pt
    \begin{subfigure}[t]{0.21\columnwidth}
        \vskip 0pt
        \centering
        \includegraphics[width=\textwidth]{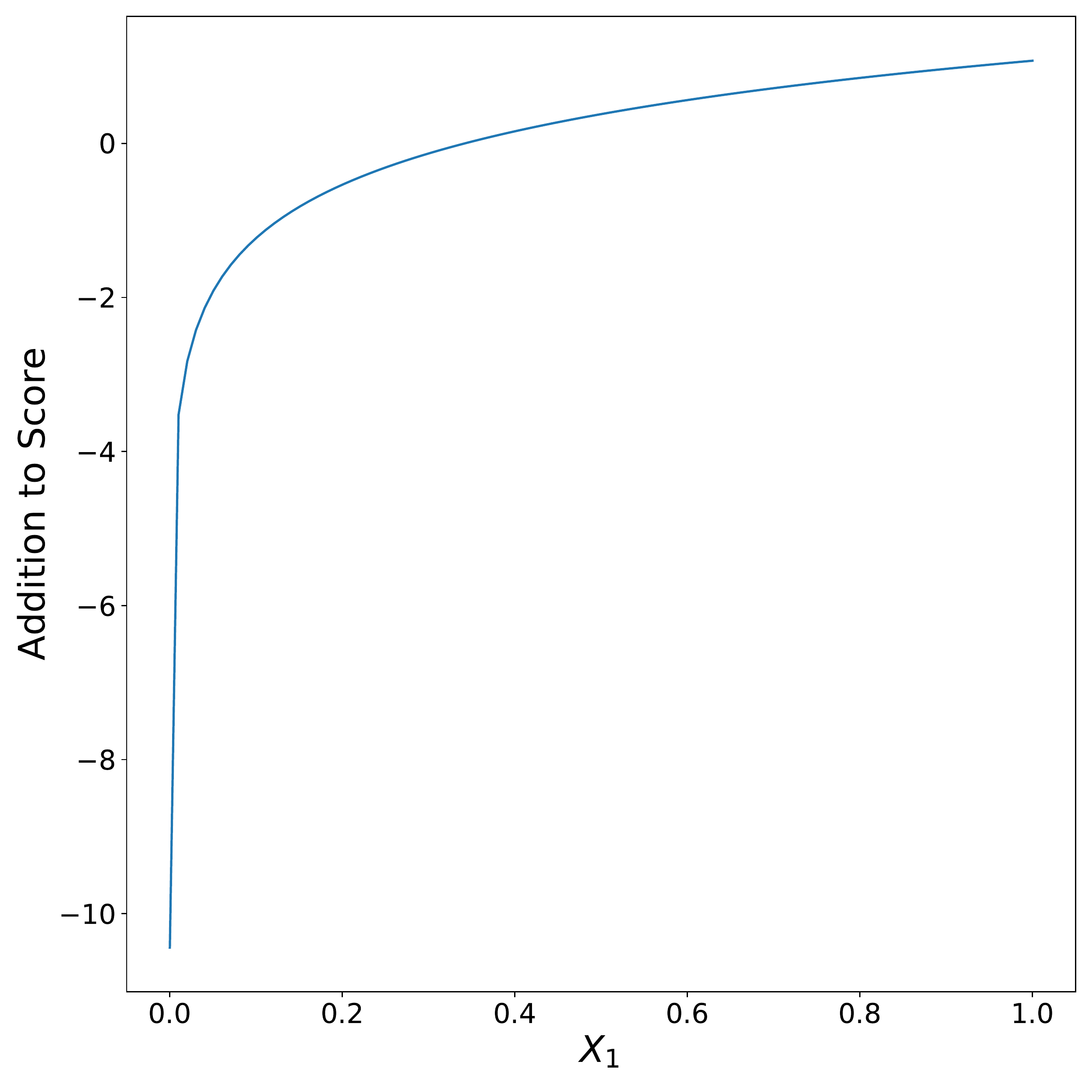}
        \caption{$\lambda = 0.0$}
    \end{subfigure}
    ~
    \begin{subfigure}[t]{0.21\columnwidth}
        \vskip 0pt
        \centering
        \includegraphics[width=\textwidth]{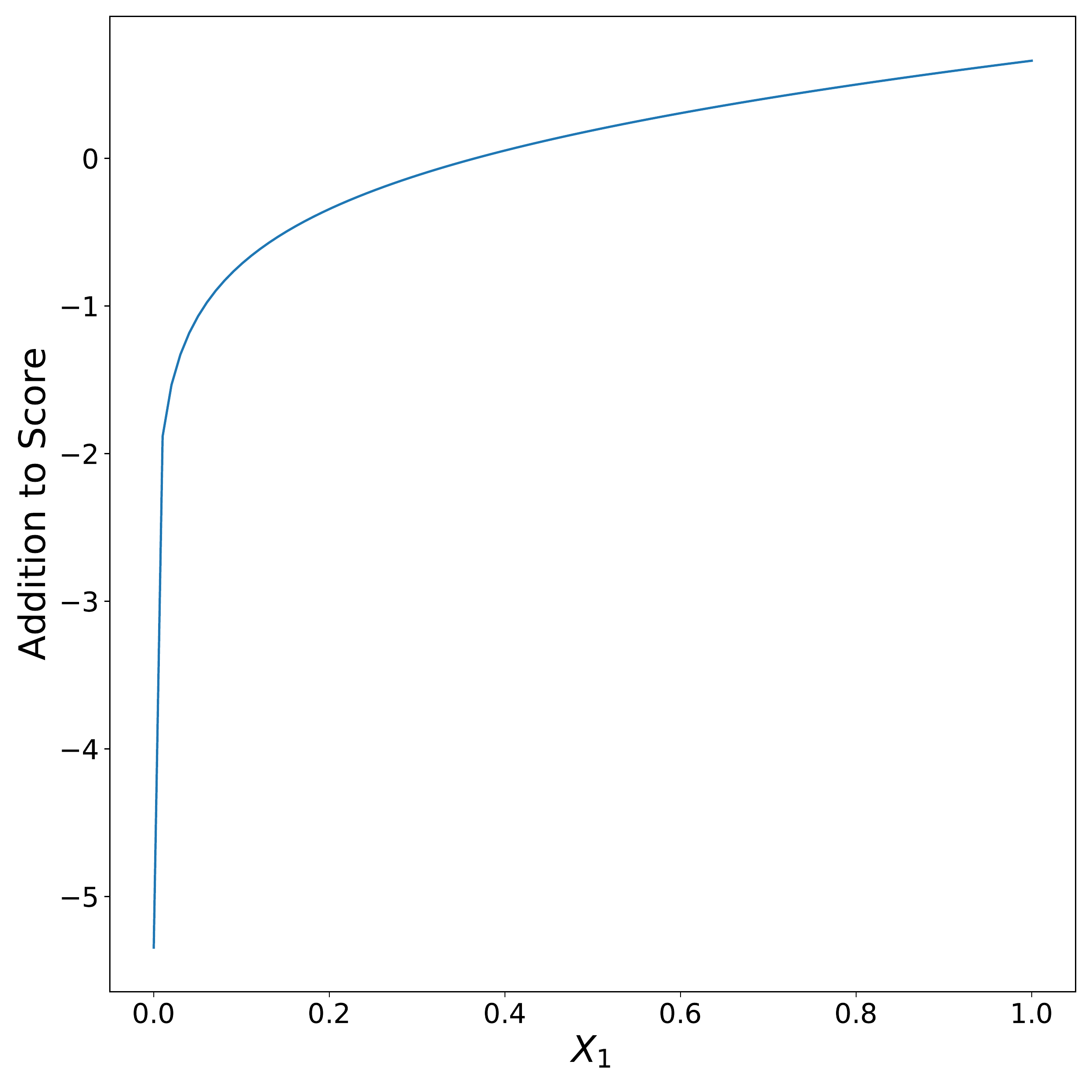}
        \caption{$\lambda = 0.5$}
    \end{subfigure}
    ~
    \begin{subfigure}[t]{0.21\columnwidth}
        \vskip 0pt
        \centering
        \includegraphics[width=\textwidth]{figs/log/new/main_effect_x1_lam_1000.pdf}
        \caption{$\lambda = 1.0$}
    \end{subfigure}
    \caption{Pure Main Effects of the log model \eqref{eq:log_model}. The purification algorithm recovers the transition from logarithmic to linear main effects.}
    \label{fig:log_mains}
\end{figure}

As shown in Fig.~\ref{fig:log_mains}, the main effects recovered at $\lambda=0$ are logarithmic, and the interaction effect completely disappears. 
As $\lambda$ increases, the main effects transition toward linearity, and the interaction effect returns, becoming the continuous variant of $XOR$. 
In fact, for any $\lambda \neq 0$, the pattern of the pure interaction effect is the same, and the overall strength of the interaction effect increases as $\lambda \rightarrow 1$.\footnote{The base of the logarithm does not alter the shape of the interaction, but does affect how rapidly $C$ varies with $\lambda$.}

\subsection{Purification Explains the Results of Prior Work}
\label{sec:discussion:explains_prior}
Our work suggests that interactions can properly be understood only after purification. 
This problem has confounded prior work studying the ability of machine learning to identify interactions. 
For example, \cite{wright2016little} studied the ability of random forests to learn interaction effects. 
The authors generated data from five different data generators designed to reflect interactions of genetic single nucleotide polymorphisms (SNPs). 
Their results appeared to lead to a pessimistic conclusion that random forests are not adept at learning interaction effects. 

\begin{table}[htb]
    \centering
    \parbox{0.45\textwidth}{
    \begin{tabular}{l|c|c|c}
    Model & SNP1 & SNP2 & SNP1xSNP2 \\
    \toprule
    Interaction Only & $0$ & $0$ & $1$ \\
    Modifier SNP & $0$ & $1$ & $1$ \\
    No Interaction & $1$ & $1$ & $0$ \\
    Redundant & $1$ & $1$ & $-1$ \\
    Synergistic & $1$ & $1$ & $1$
    \end{tabular}
    \caption{Data Generator Coefficients, Unpurified \label{tab:data_generators_raw}}
    }
    ~
    \parbox{0.45\textwidth}{
    \begin{tabular}{l|c|c|c}
    Model & SNP1 & SNP2 & SNP1$\lxor$SNP2 \\
    \toprule
    Interaction Only & $0.5$ & $0.5$ & $0.25$ \\
    Modifier SNP & $0.5$ & $1.5$ & $0.25$ \\
    No Interaction & $1$ & $1$ & $0$ \\
    Redundant & $0.5$ & $0.5$ & $-0.25$ \\
    Synergistic & $1.5$ & $1.5$ & $0.25$
    \end{tabular}
    \caption{Data Generator Coefficients, Purified\label{tab:data_generators_pure}}
    }
\end{table}

\input{wright_figure_arxiv.tex}

However, as shown in Tables~\ref{tab:data_generators_raw},\ref{tab:data_generators_pure} (and visualized in Fig.~\ref{fig:wright} of the Supplement), these data generators look \emph{very} different before and after purification.
In particular, the data generation schemes differ dramatically in the strength of the pure main effects: the ``synergistic" model has main effects three times as strong as the main effects in the ``interaction only" model. 
In contrast, the interaction effect is the same strength for all data generation schemes (except for the ``No Interaction" setting). 
In light of the purified data generators, the results of Figs.~3 and 4 of \cite{wright2016little} suggest a more optimistic conclusion: in the case of interaction effects of equal strength, the random forest preferentially recovers interactions of variables with stronger main effects. 
This result underscores the necessity of purifying data generation schemes before studying the inductive biases of machine learning models.

\subsection{Purification Reveals Noisy Estimates}
\label{sec:calculating:sketching}

As we saw in experiments (Section~\ref{sec:experiments}), purification can produce main effects that are less smooth than the main effects of the original model. 
Purification does not change the model, so 
the mass that made a main effect less smooth was hidden in the interactions prior to mass moving. 
High variance in terms can hurt interpretability and likely should be regularized out for more robust estimation. 

We are currently investigating post-hoc regularization methods that simplify the main effects to reduce the variance induced when estimating interaction effects, and revealed by the purification process.  Preliminary results suggest that main effects often can be simplified, which not only makes them easier to interpret, but, in some cases, makes them modestly more accurate on test data.

\section{Conclusions and Future Work}
\label{sec:conclusions}
We have shown that the non-identifiability of interaction effects in additive models is a problem for model interpretability -- equivalent models can produce contradictory interpretations.  
We have proposed to use the fANOVA decomposition to recover meaningful interaction effects, and have presented an efficient algorithm to exactly recover this decomposition for piecewise-constant functions such as tree-based estimators. 
In the past, algorithms such as GA2M have been designed to prioritize main effects over interactions during estimation; our method of post-hoc purification returns an identifiable form of any tree-based model, and thus frees model designers to separate estimation procedures from purification procedures. 
Finally, we have applied this approach to learn pure interaction effects from several datasets, and seen that the interpretation of these effects changes in response to the data distribution. 
This underscores the importance of specifying the data distribution before attempting to interpret any estimated effects. 
The true density $p(x)$ is the correct data distribution for model interpretation, but $p(x)$ is seldom known, so we are interested in future work to improve estimators of $p(x)$ in order to improve model interpretability.

\subsection*{Acknowledgements}
This work was created during an internship at Microsoft Research, and completed under the support of a fellowship from the Center for Machine Learning and Healthcare.

We would like to thank the anonymous reviewers, whose feedback greatly assisted in the presentation of this work.

\bibliographystyle{abbrvnat}
\bibliography{additive_models}

\clearpage
\appendix

\section{Analysis}
\label{sec:appendix:proof}

We will prove the rate of convergence by setting an upper-bound on $M^t$:
\begin{lemma}
For any $T_{a,b}, w, \Omega$, if iteration $t$ set the column means to be zero, then \label{thm:m_t1}
\begin{align}
    M^{t+1} \leq &= \sum_{j \in \Omega_b}\min_{\phi_j}\abs{\sum_{i \in \Omega_a}(\frac{w_{i,j}}{w_{i, \cdot}} - \phi_j)\big(\min_{\psi_i}\sum_{k \in \Omega_b}(w_{i,k} - \psi_i w_{\cdot, k})c_k^{t-1}\big)}
\end{align}
\end{lemma}

\subsection{Proof of Lemma \ref{thm:m_t1}}
\begin{proof}
Let us consider a matrix $T$ representing the interaction effect of variables $X_a$ and $X_b$. 
Let $X_a$ take on values from the set $\Omega_a,$, and $X_b$ take on values from the set $\Omega_b$, with $m=|\Omega_a|$, $n=|\Omega_b|$. 
Let $w$ be a density defined on $\Omega_a$ and $\Omega_b$, normalized such that $\sum_{i\in \Omega_a}\sum_{j \in \Omega_b}w_{i,j} = 1$. 
Without loss of generality, we assume that $T$ is mean-centered such that $\sum_{i\in \Omega_a}\sum_{j \in \Omega_b}w_{i,j}T_{i,j} = 0$. 
For clarity, we also use the shorthand:
\begin{subequations}
\begin{align}
    w_{\cdot, j} &= \sum_{i \in \Omega_a} w_{i, j} \\
    w_{i, \cdot} &= \sum_{j \in \Omega_b} w_{i, j}
\end{align}
\end{subequations}
\noindent
Without loss of generality, we can assume that iteration $t$ set all of the column-means to zero. Then iteration $t+1$ will set the row means to zero, and: 
\begin{subequations}
\begin{align}
    M^{t+1} &= \sum_{j \in \Omega_b}w_{\cdot, j}\abs{c_j^{t+1}} \\
    &= \sum_{j \in \Omega_b}w_{\cdot, j}\abs{\frac{1}{w_{\cdot, j}}\sum_{i \in \Omega_a}w_{i,j}r_i^t} \\
    &= \sum_{j \in \Omega_b}\abs{\sum_{i \in \Omega_a}w_{i,j}r_i^t} \\
    &= \sum_{j \in \Omega_b}\abs{\sum_{i \in \Omega_a}w_{i,j}\frac{1}{w_{i, \cdot}}\big(\sum_{k \in \Omega_b}w_{i,k}c_k^{t-1}\big)} \\
    &= \sum_{j \in \Omega_b}\min_{\phi_j}\abs{\sum_{i \in \Omega_a}(w_{i,j} - \phi_j w_{i, \cdot})\frac{1}{w_{i, \cdot}}\big(\sum_{k \in \Omega_b}w_{i,k}c_k^{t-1}\big)} \label{eq:intro_min} \\
    &= \sum_{j \in \Omega_b}\min_{\phi_j}\abs{\sum_{i \in \Omega_a}(\frac{w_{i,j}}{w_{i, \cdot}} - \phi_j)\big(\sum_{k \in \Omega_b}w_{i,k}c_k^{t-1}\big)} \\
    &= \sum_{j \in \Omega_b}\min_{\phi_j}\abs{\sum_{i \in \Omega_a}(\frac{w_{i,j}}{w_{i, \cdot}} - \phi_j)\big(\min_{\psi_i}\sum_{k \in \Omega_b}(w_{i,k} - \psi_i w_{\cdot, k})c_k^{t-1}\big)}
\end{align}
\end{subequations}
where \eqref{eq:intro_min} holds because $\sum_{i \in \Omega_a}\sum_{k \in \Omega_b} w_{i,k}c_k^{t-1}$ is the overall mean of the matrix, which is zero.
\end{proof}

\subsection{Proof of Theorem \ref{thm:uniform}}
\begin{proof}
Under the assumptions of Lemma~\ref{thm:m_t1}, 
\begin{subequations}
\begin{align}
    M^{t+1} &\leq \sum_{j \in \Omega_b}\min_{\phi_j}\abs{\sum_{i \in \Omega_a}(\frac{w_{i,j}}{w_{i, \cdot}} - \phi_j)\big(\min_{\psi_i}\sum_{k \in \Omega_b}(w_{i,k} - \psi_i w_{\cdot, k})c_k^{t-1}\big)} & \text{by Lemma \ref{thm:m_t1}} \\
    &= \sum_{j \in \Omega_b}\abs{\sum_{i \in \Omega_a}(\frac{w_{i,j}}{w_{i, \cdot}} - \frac{1}{n})\big(\min_{\psi_i}\sum_{k \in \Omega_b}(w_{i,k} - \psi_i w_{\cdot, k})c_k^{t-1}\big)} \\
    &= \sum_{j \in \Omega_b}\abs{\sum_{i \in \Omega_a}(0)\big(\min_{\psi_i}\sum_{k \in \Omega_b}(w_{i,k} - \psi_i w_{\cdot, k})c_k^{t-1}\big)} & \text{for uniform }w\\
    &= 0
\end{align}
\end{subequations}
\end{proof}

\subsection{Proof of Theorem \ref{thm:convergence}}
\begin{proof}
For any normalized $w$,
\begin{subequations}
\begin{align}
    M^{t+1} &= \sum_{j \in \Omega_b}\min_{\phi_j}\abs{\sum_{i \in \Omega_a}(\frac{w_{i,j}}{w_{i, \cdot}} - \phi_j)\big(\min_{\psi_{i,j}}\sum_{k \in \Omega_b}(w_{i,k} - \psi_{i,j} w_{\cdot, k})c_k^{t-1}\big)} & \text{by Lemma \ref{thm:m_t1}}\\
    &= \sum_{j \in \Omega_b}\min_{\phi_j}\abs{\sum_{i \in \Omega_a}(\frac{w_{i,j}}{w_{i, \cdot}} - \phi_j)\min_{\psi_{i,j}}w_{i, \cdot}\sum_{k \in \Omega_b}(\frac{w_{i,k}}{w_{i, \cdot}} - \frac{\psi_{i,j}}{w_{i, \cdot}} w_{\cdot, k})c_k^{t-1}} \\
    &= \sum_{j \in \Omega_b}\min_{\phi_j}\abs{\sum_{i \in \Omega_a}(\frac{w_{i,j}}{w_{i, \cdot}} - \phi_j)\sum_{k \in \Omega_b}\sum_{l \neq i} w_{l,k}c_k^{t-1}} \\
    &\leq \sum_{j \in \Omega_b}\min_{\phi_j}\abs{\sum_{i \in \Omega_a}(\frac{w_{i,j}}{w_{i, \cdot}} - \phi_j)\sum_{k \in \Omega_b}w_{\cdot,k}c_k^{t-1}} \\
    &\leq \sum_{j \in \Omega_b}\min_{\phi_j}\abs{\sum_{i \in \Omega_a}(\frac{w_{i,j}}{w_{i, \cdot}} - \phi_j)\sum_{k \in K^{+}}w_{\cdot,k}c_k^{t-1}} & \quad K^{+} = \{ k \in \Omega_b: c_{k}^{t-1} > 0 \} \\
    &\leq \sum_{j \in \Omega_b}\min_{\phi_j}\abs{\sum_{i \in \Omega_a}(\frac{w_{i,j}}{w_{i, \cdot}} - \phi_j)\frac{1}{2}M^{t-1}} \\
    &\leq \frac{1}{2}M^{t-1}\sum_{j \in \Omega_b}\min_{\phi_j}\abs{\sum_{i \in \Omega_a}(\frac{w_{i,j}}{w_{i, \cdot}} - \phi_j)} \\
    &\leq \frac{1}{2}M^{t-1}\sum_{i \in \Omega_a}\sum_{j \in \Omega_b}\min_{\phi_j}\abs{\frac{w_{i,j}}{w_{i, \cdot}} - \phi_j} \\
    &\leq \frac{1}{2}M^{t-1}\sum_{i \in \Omega_a}w_{i, \cdot} \\
    &= \frac{1}{2}M^{t-1}
\end{align}
\end{subequations}
So the divergence from the fANOVA decomposition is cut in half each iteration. 
This is a loose bound which could be tightened by examining the dispersion of the density.
\end{proof}

\clearpage
\section{Empirically Measuring Purification Convergence}
\label{sec:appendix:convergence}
Here we show a variety of convergence plots for Algorithm~\ref{alg:helper} on data generated according to the setups described in Section~\ref{sec:alg:measuring}. 

\begin{figure}[ht]
    \centering
    \begin{subfigure}[t]{0.31\columnwidth}
        \vskip 0pt
        \centering
        \includegraphics[width=\textwidth]{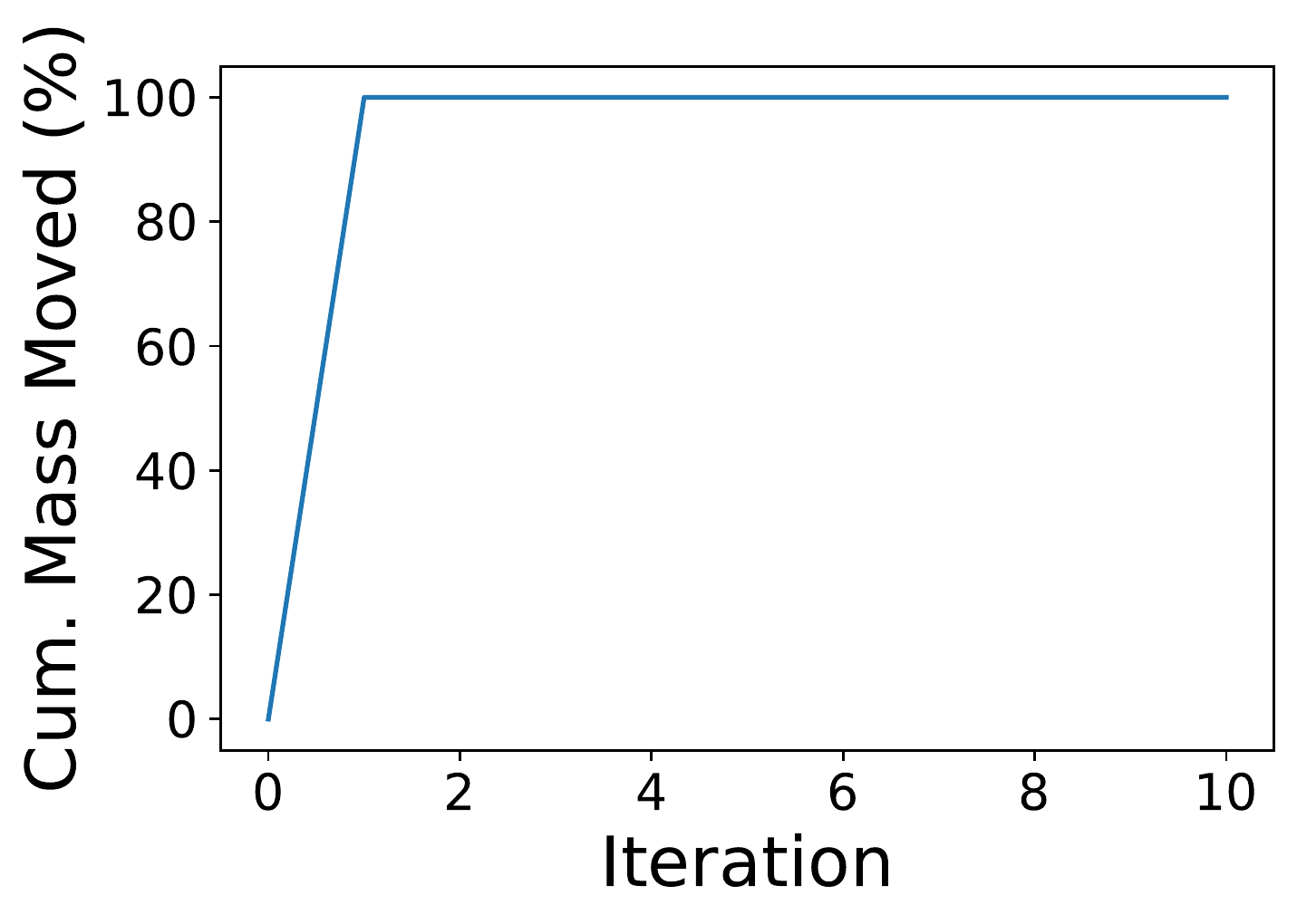}
        \caption{$\sigma=1$, $P=2$}
    \end{subfigure}
    ~
    \begin{subfigure}[t]{0.31\columnwidth}
        \vskip 0pt
        \centering
        \includegraphics[width=\textwidth]{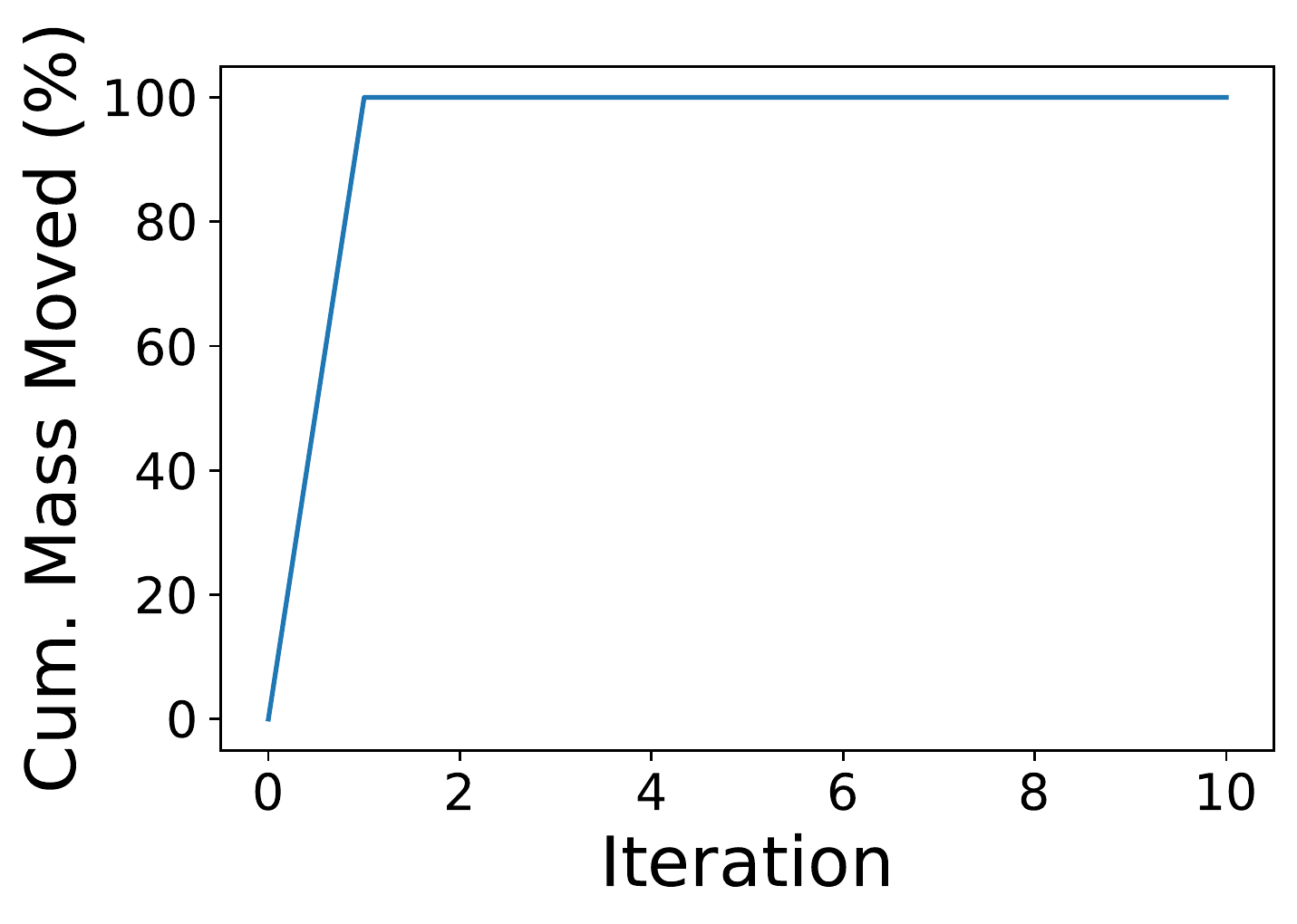}
        \caption{$\sigma=1$, $P=25$}
    \end{subfigure}
    ~
    \begin{subfigure}[t]{0.31\columnwidth}
        \vskip 0pt
        \centering
        \includegraphics[width=\textwidth]{figs/convergence/unif_sigma_1_ndims_100.pdf}
        \caption{$\sigma=1$, $P=100$}
    \end{subfigure}
    ~
    \begin{subfigure}[t]{0.31\columnwidth}
        \vskip 0pt
        \centering
        \includegraphics[width=\textwidth]{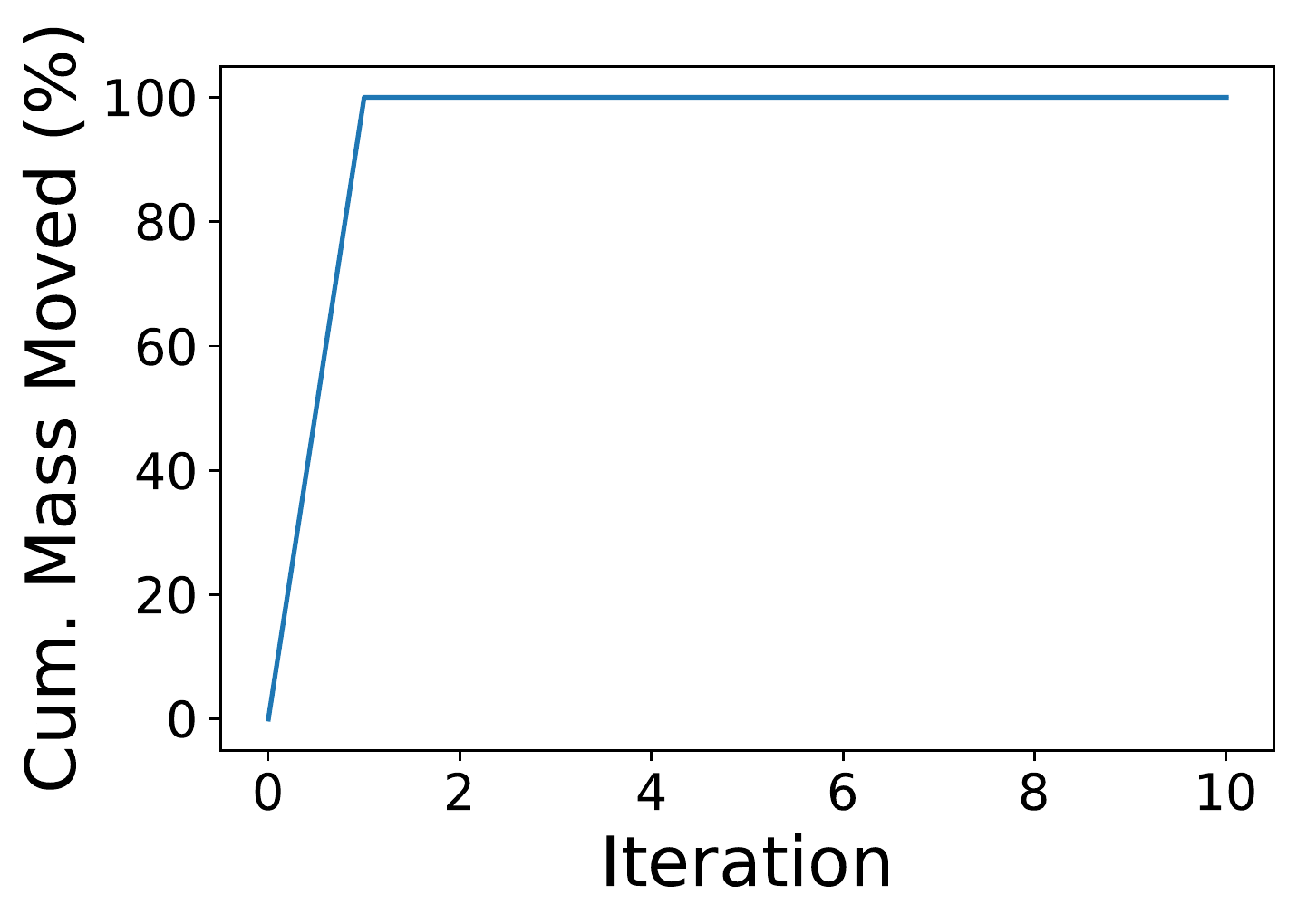}
        \caption{$\sigma=10$, $P=2$}
    \end{subfigure}
    ~
    \begin{subfigure}[t]{0.31\columnwidth}
        \vskip 0pt
        \centering
        \includegraphics[width=\textwidth]{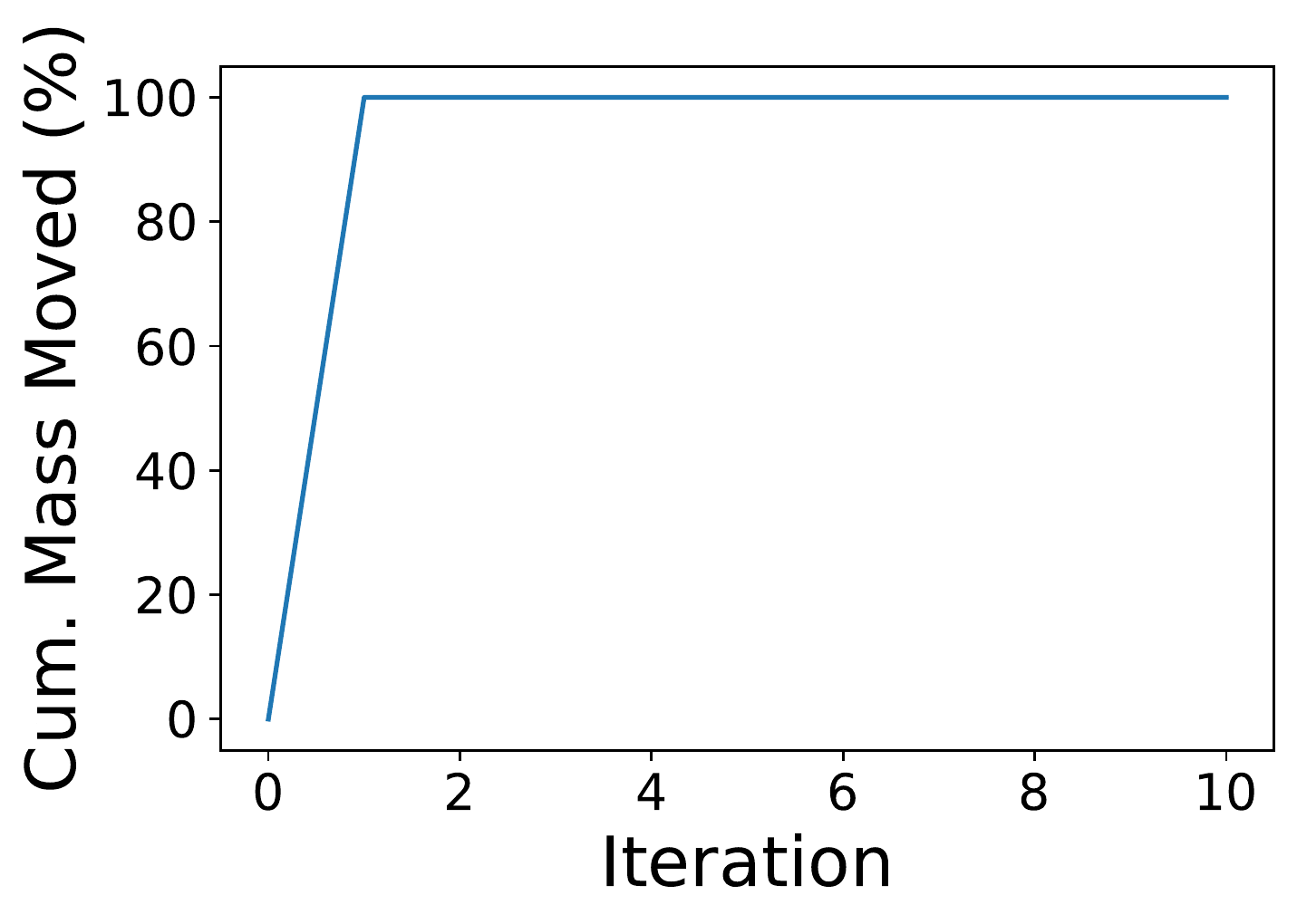}
        \caption{$\sigma=10$, $P=25$}
    \end{subfigure}
    ~
    \begin{subfigure}[t]{0.31\columnwidth}
        \vskip 0pt
        \centering
        \includegraphics[width=\textwidth]{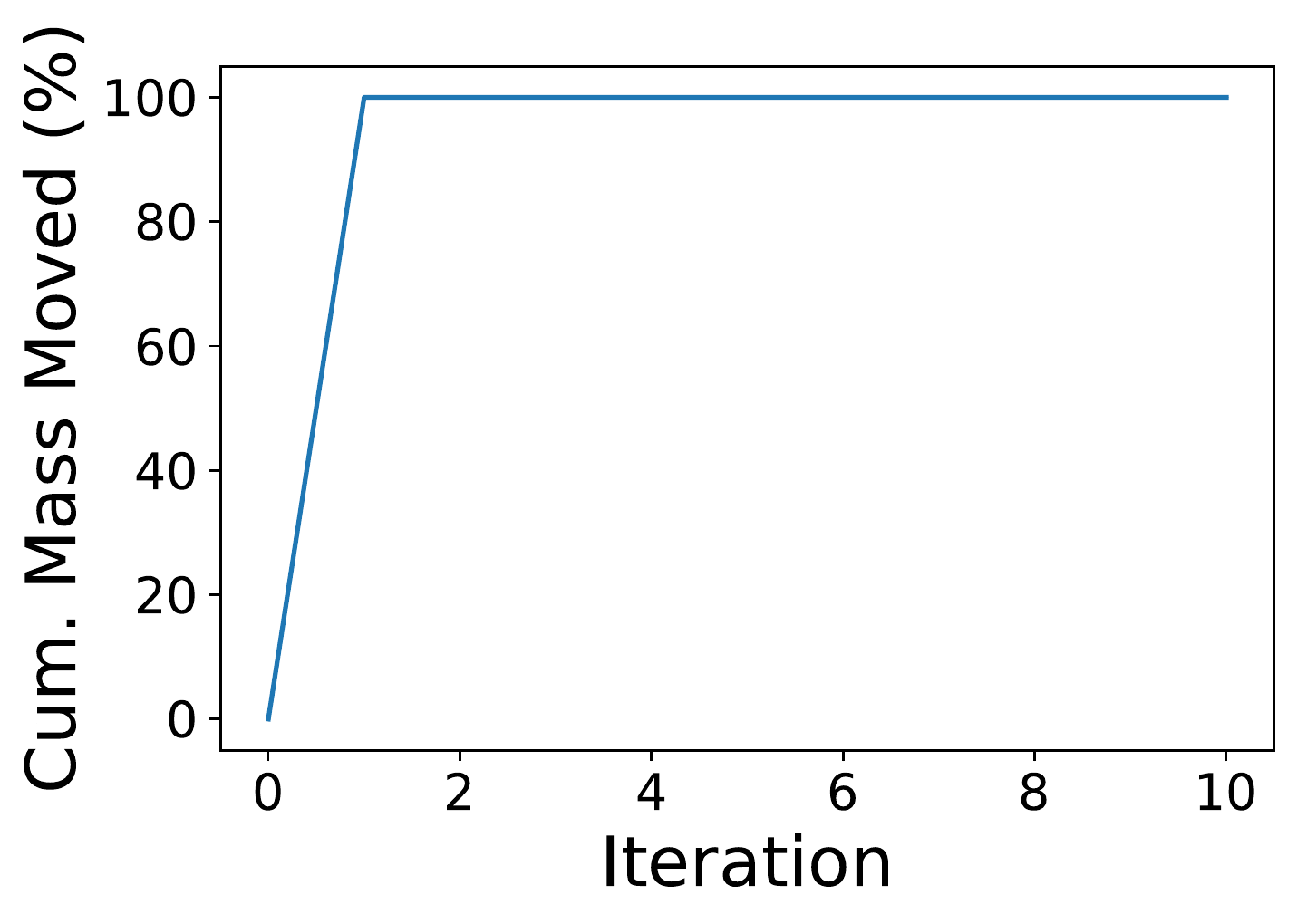}
        \caption{$\sigma=10$, $P=100$}
    \end{subfigure}
    ~
    \begin{subfigure}[t]{0.31\columnwidth}
        \vskip 0pt
        \centering
        \includegraphics[width=\textwidth]{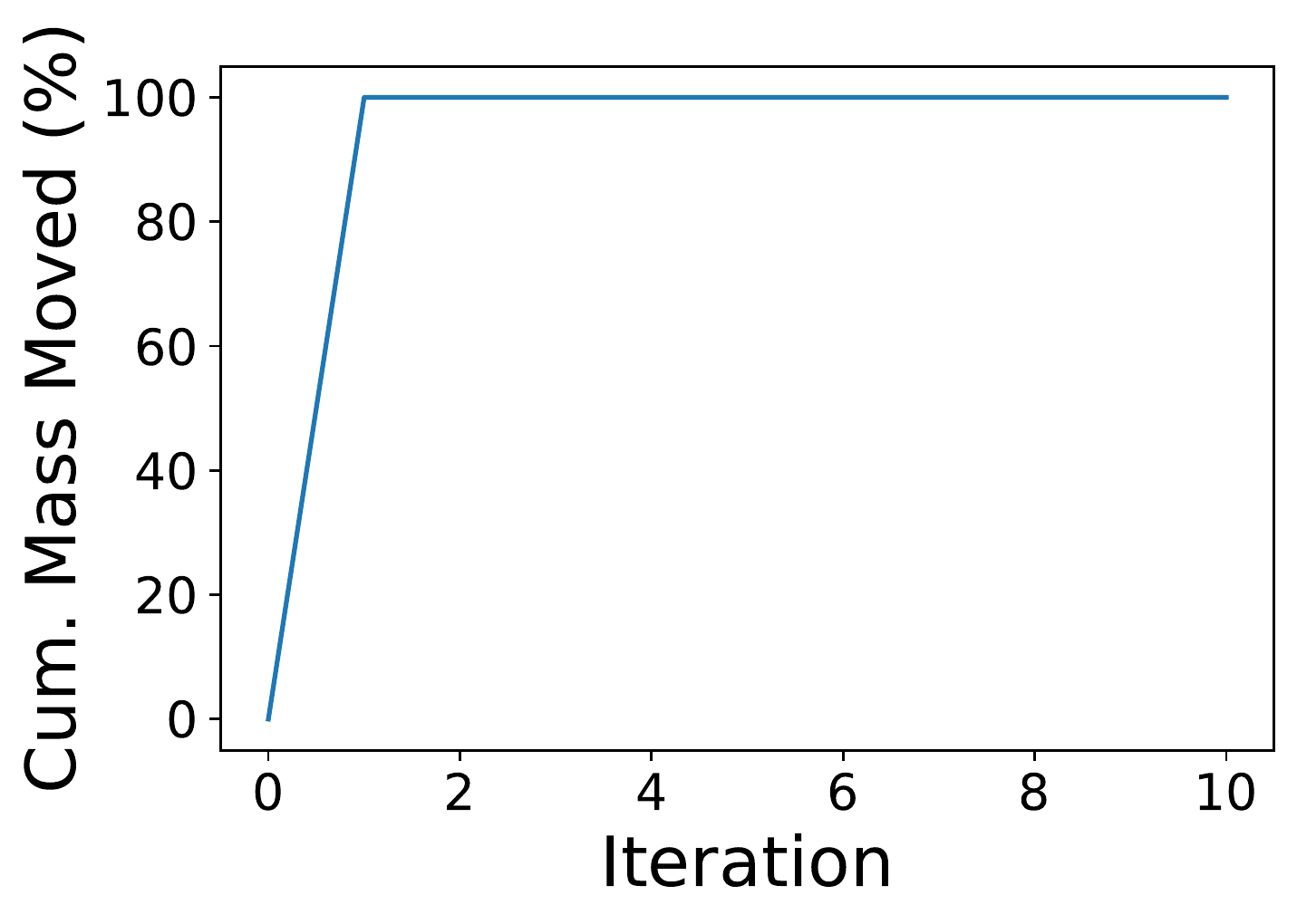}
        \caption{$\sigma=100$, $P=2$}
    \end{subfigure}
    ~
    \begin{subfigure}[t]{0.31\columnwidth}
        \vskip 0pt
        \centering
        \includegraphics[width=\textwidth]{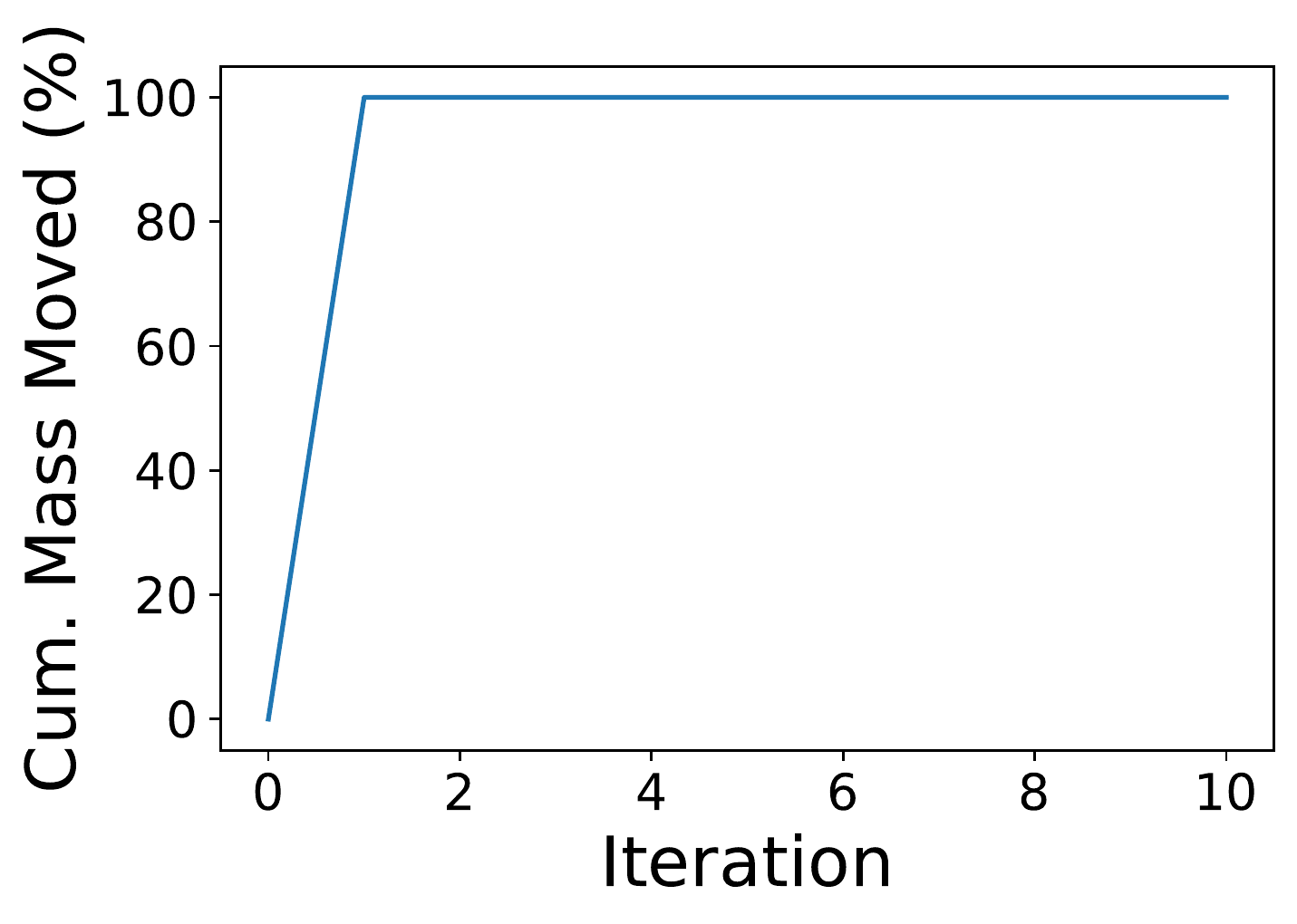}
        \caption{$\sigma=100$, $P=25$}
    \end{subfigure}
    ~
    \begin{subfigure}[t]{0.31\columnwidth}
        \vskip 0pt
        \centering
        \includegraphics[width=\textwidth]{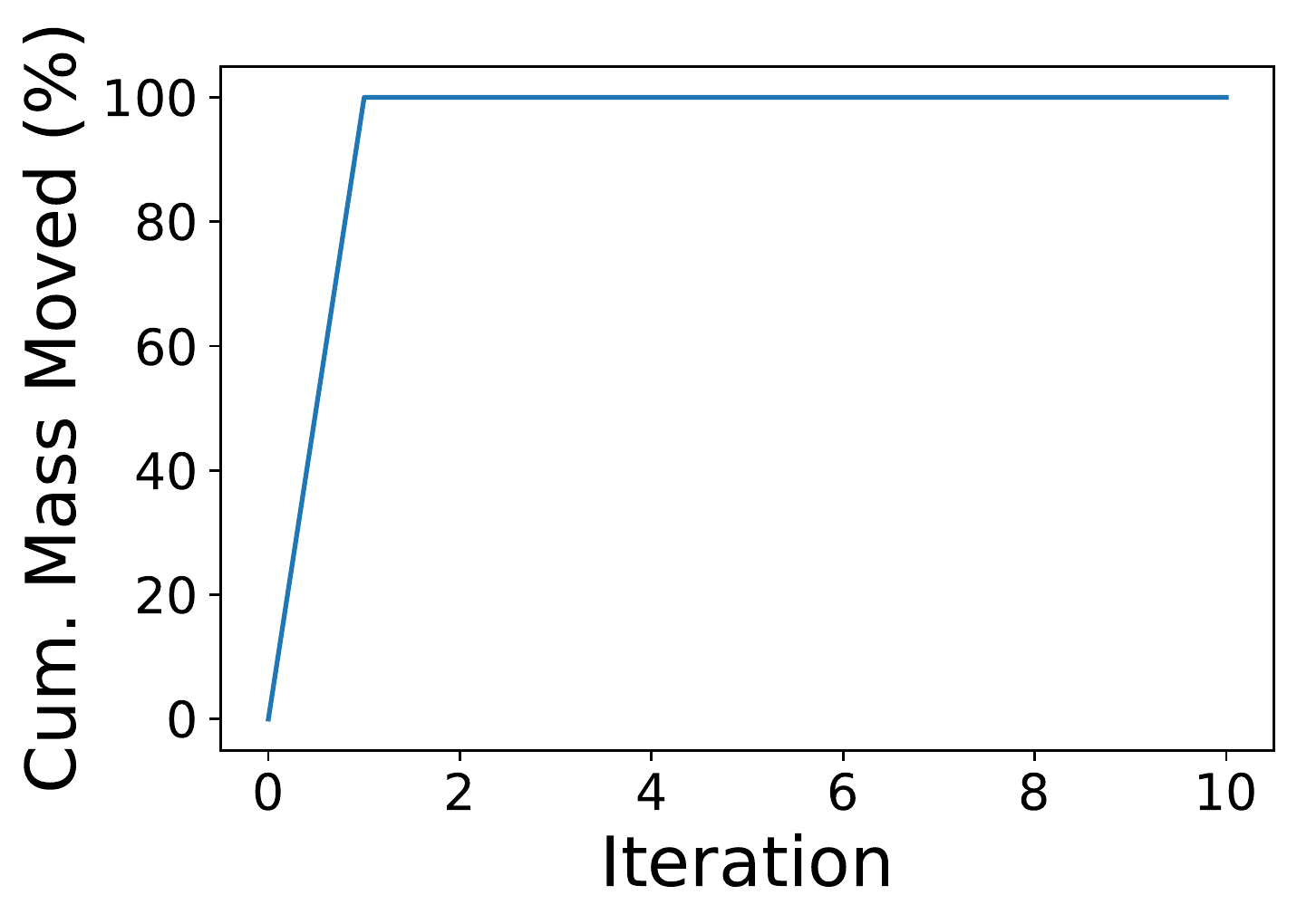}
        \caption{$\sigma=100$, $P=100$}
    \end{subfigure}
    \caption{Convergence of the mass-moving algorithm under uniform weight distributions. In all settings, the unpurified effect matrix is drawn from $N(0, \sigma I)$ of dimension $P$ while the weighting is uniform. In all settings, the algorithm converges in a single iteration. \label{fig:convergence_uniform}}
\end{figure}

\begin{figure}[ht]
    \centering
    \begin{subfigure}[t]{0.31\columnwidth}
        \vskip 0pt
        \centering
        \includegraphics[width=\textwidth]{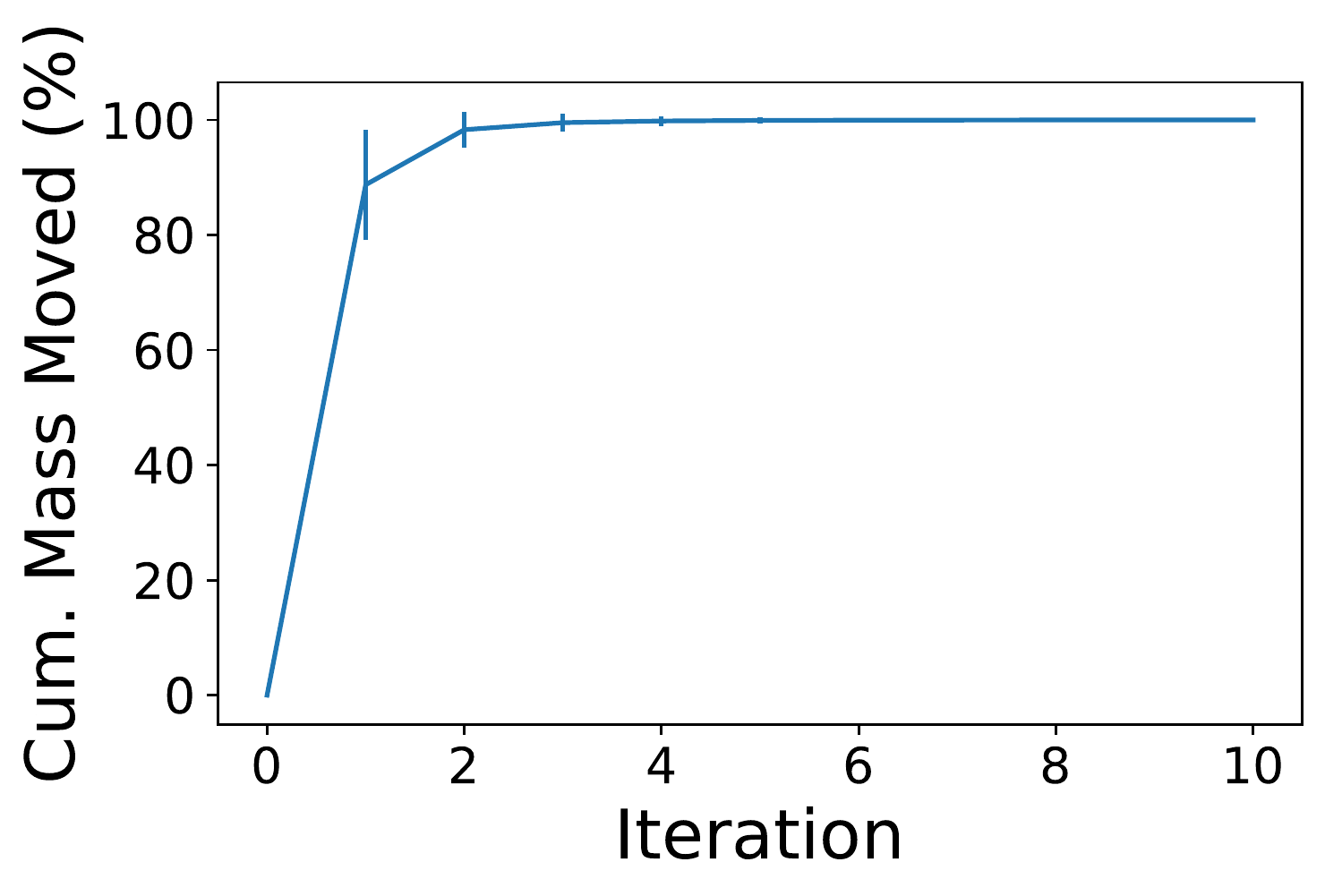}
        \caption{$\sigma=1$, $P=2$}
    \end{subfigure}
    ~
    \begin{subfigure}[t]{0.31\columnwidth}
        \vskip 0pt
        \centering
        \includegraphics[width=\textwidth]{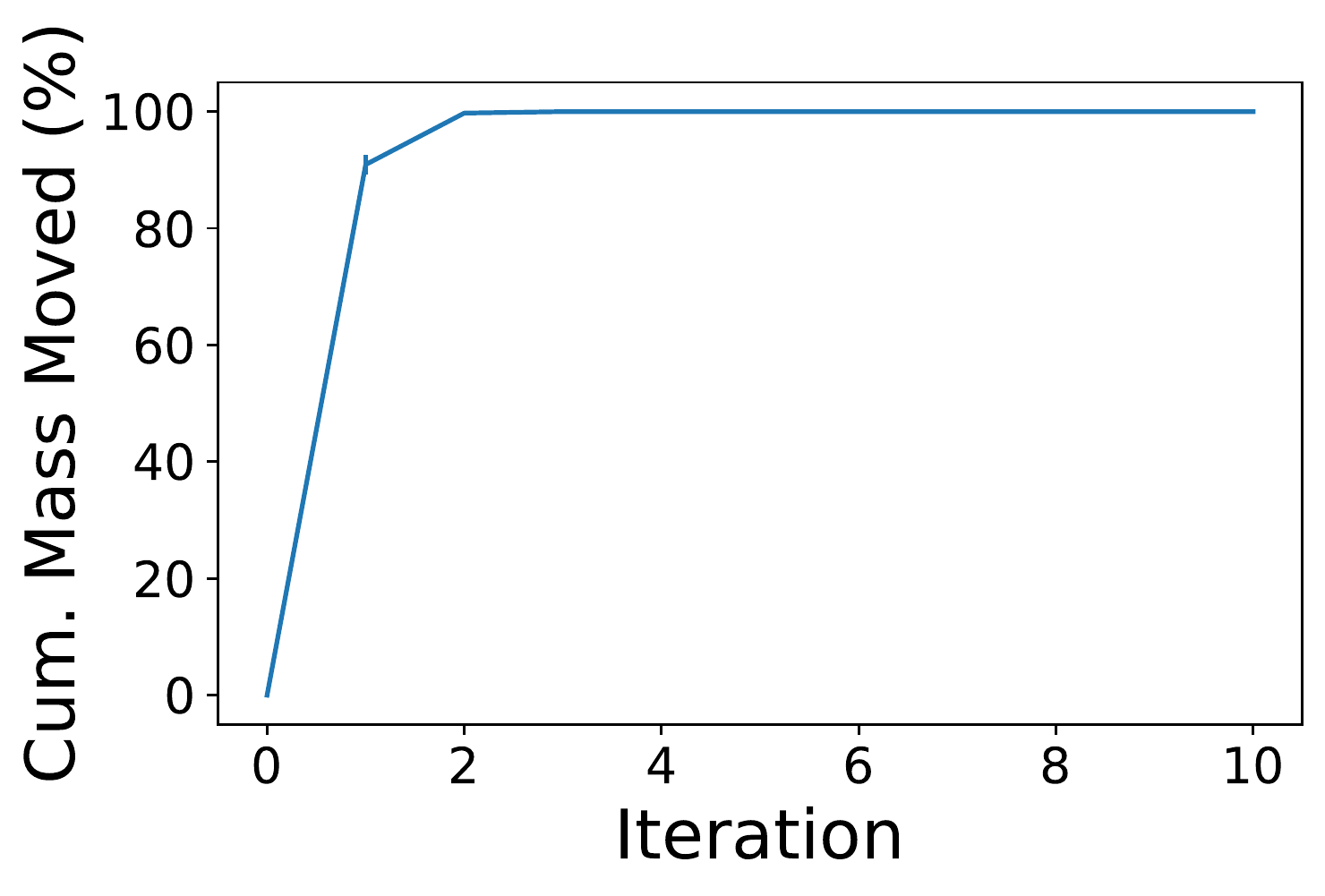}
        \caption{$\sigma=1$, $P=25$}
    \end{subfigure}
    ~
    \begin{subfigure}[t]{0.31\columnwidth}
        \vskip 0pt
        \centering
        \includegraphics[width=\textwidth]{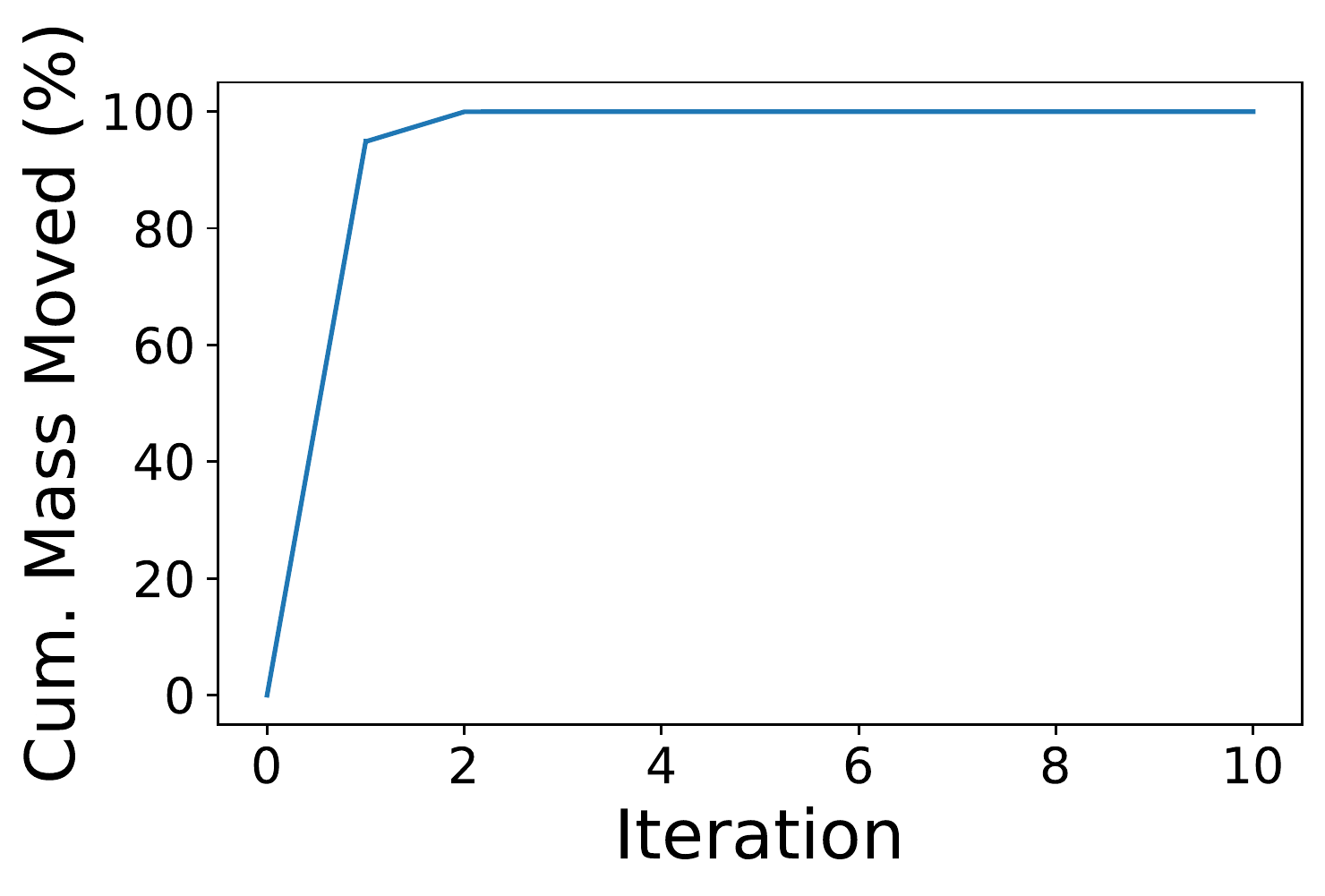}
        \caption{$\sigma=1$, $P=100$}
    \end{subfigure}
    ~
    \begin{subfigure}[t]{0.31\columnwidth}
        \vskip 0pt
        \centering
        \includegraphics[width=\textwidth]{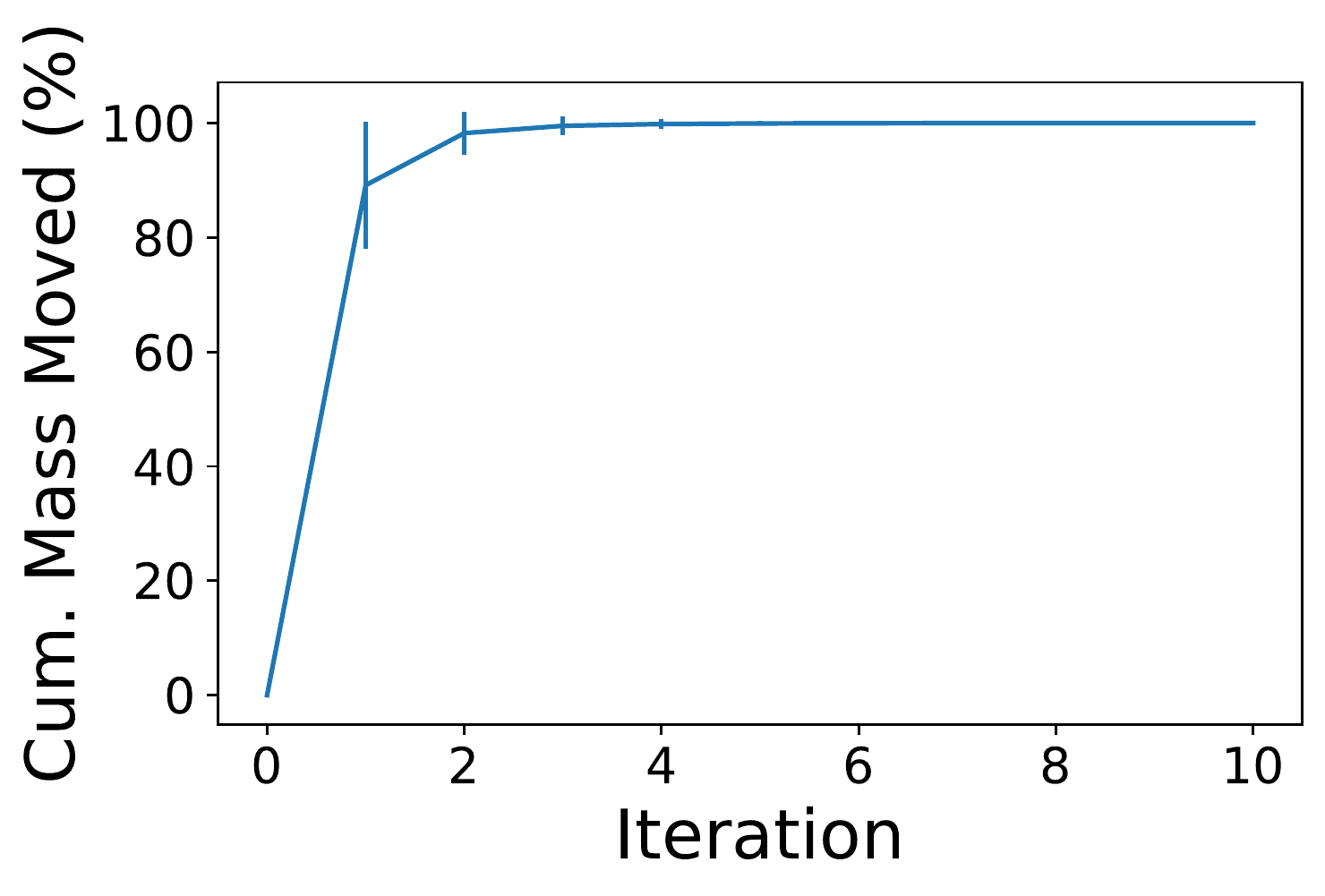}
        \caption{$\sigma=10$, $P=2$}
    \end{subfigure}
    ~
    \begin{subfigure}[t]{0.31\columnwidth}
        \vskip 0pt
        \centering
        \includegraphics[width=\textwidth]{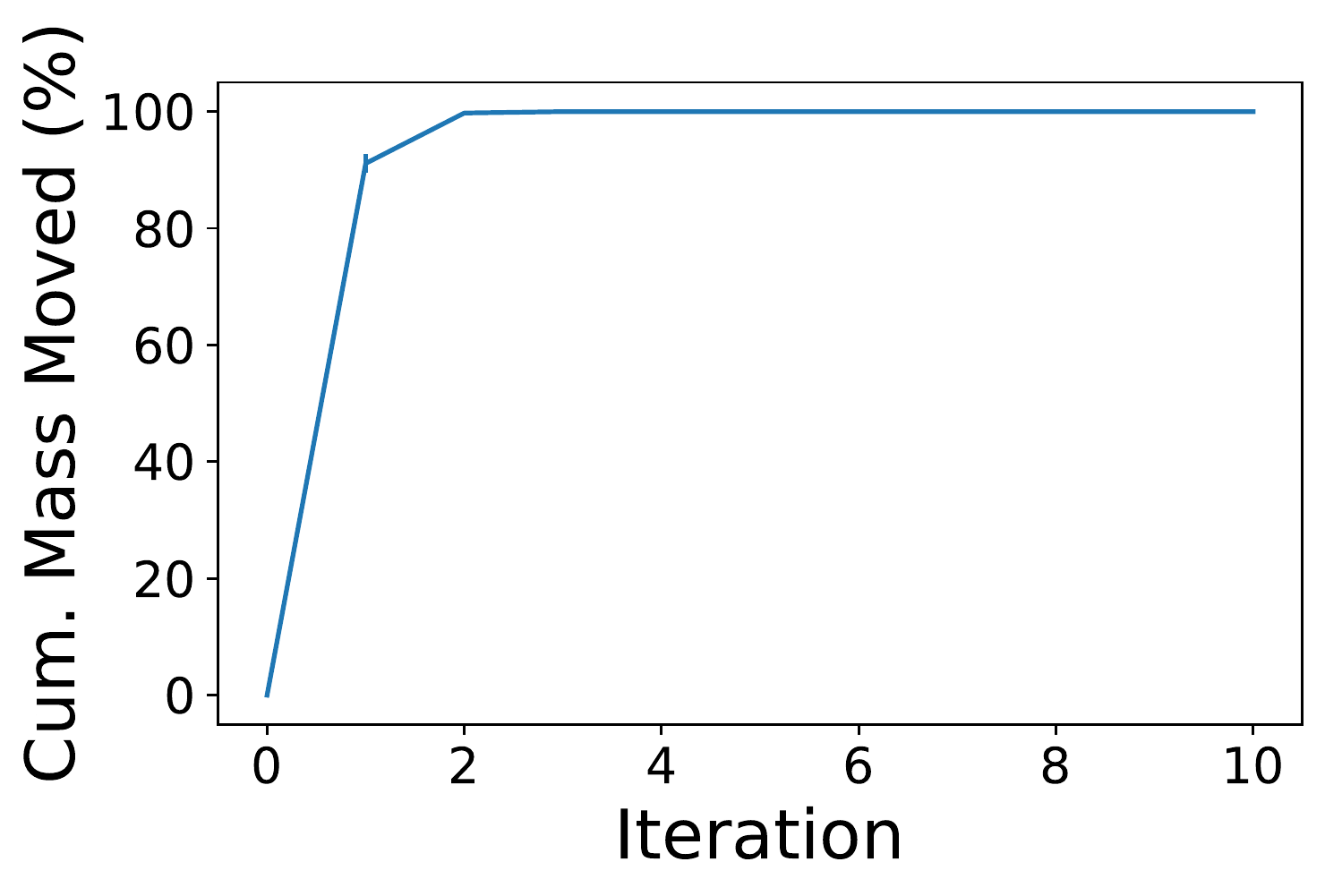}
        \caption{$\sigma=10$, $P=25$}
    \end{subfigure}
    ~
    \begin{subfigure}[t]{0.31\columnwidth}
        \vskip 0pt
        \centering
        \includegraphics[width=\textwidth]{figs/convergence/rand_sigma_10_ndims_100.pdf}
        \caption{$\sigma=10$, $P=100$}
    \end{subfigure}
    ~
    \begin{subfigure}[t]{0.31\columnwidth}
        \vskip 0pt
        \centering
        \includegraphics[width=\textwidth]{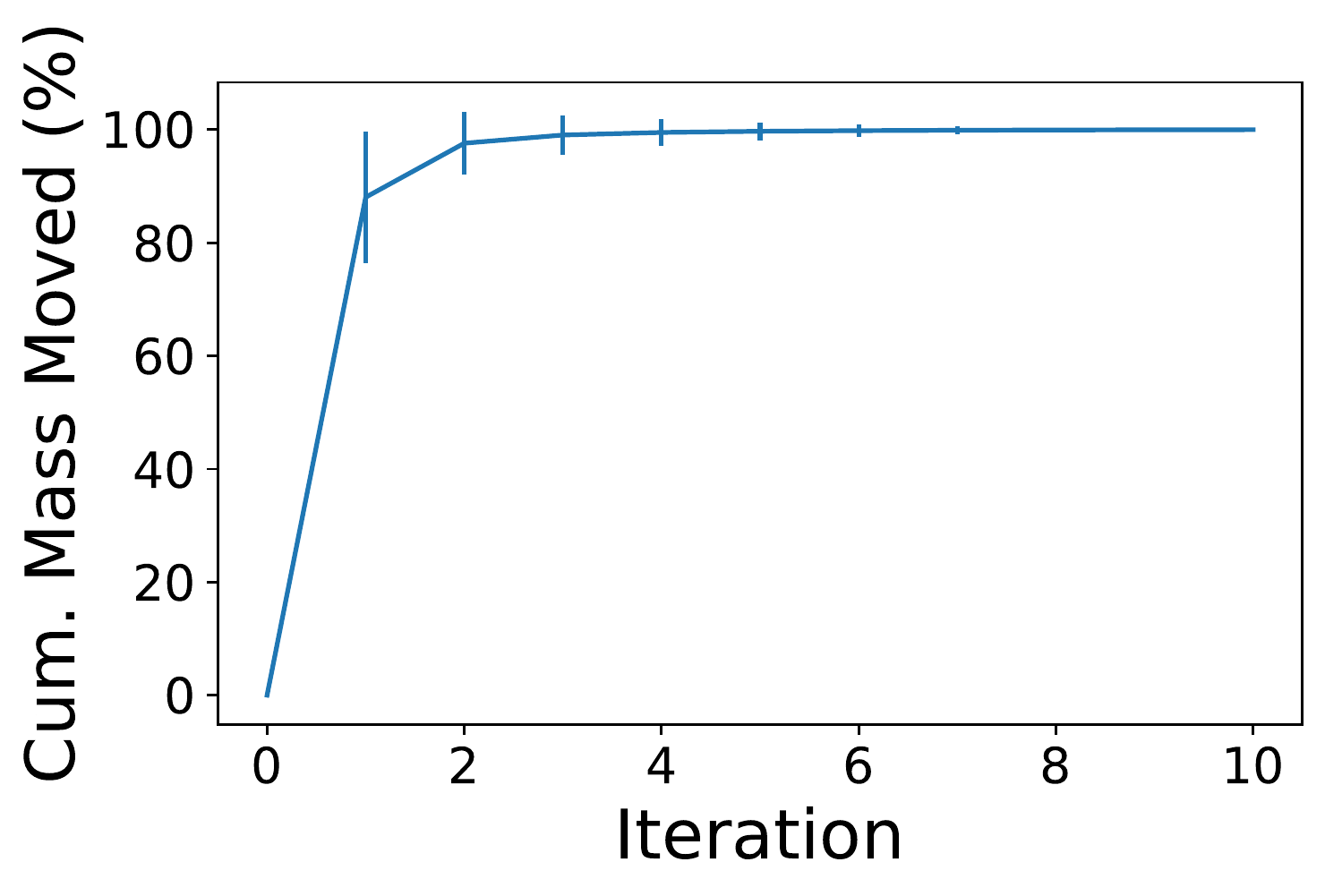}
        \caption{$\sigma=100$, $P=2$}
    \end{subfigure}
    ~
    \begin{subfigure}[t]{0.31\columnwidth}
        \vskip 0pt
        \centering
        \includegraphics[width=\textwidth]{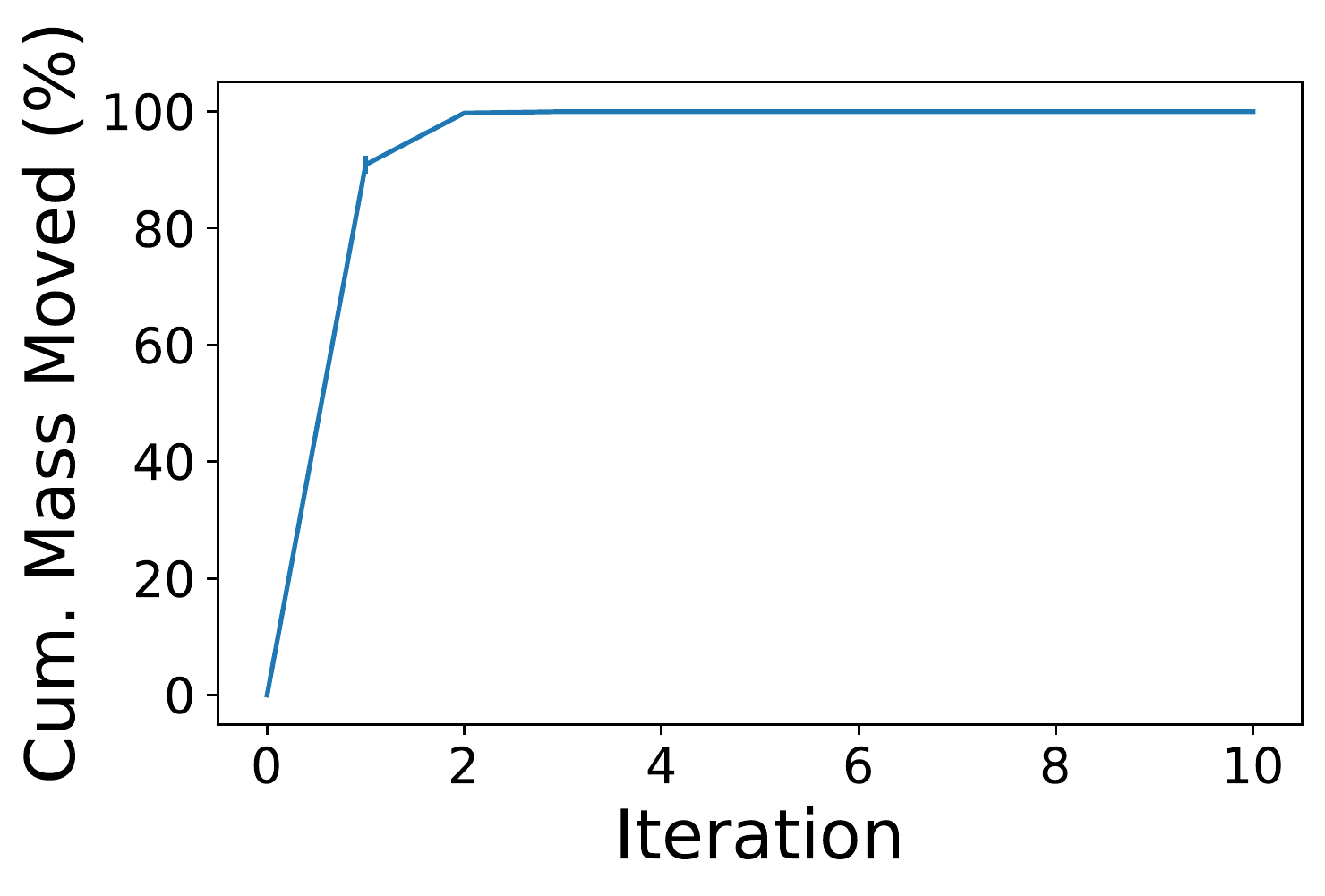}
        \caption{$\sigma=100$, $P=25$}
    \end{subfigure}
    ~
    \begin{subfigure}[t]{0.31\columnwidth}
        \vskip 0pt
        \centering
        \includegraphics[width=\textwidth]{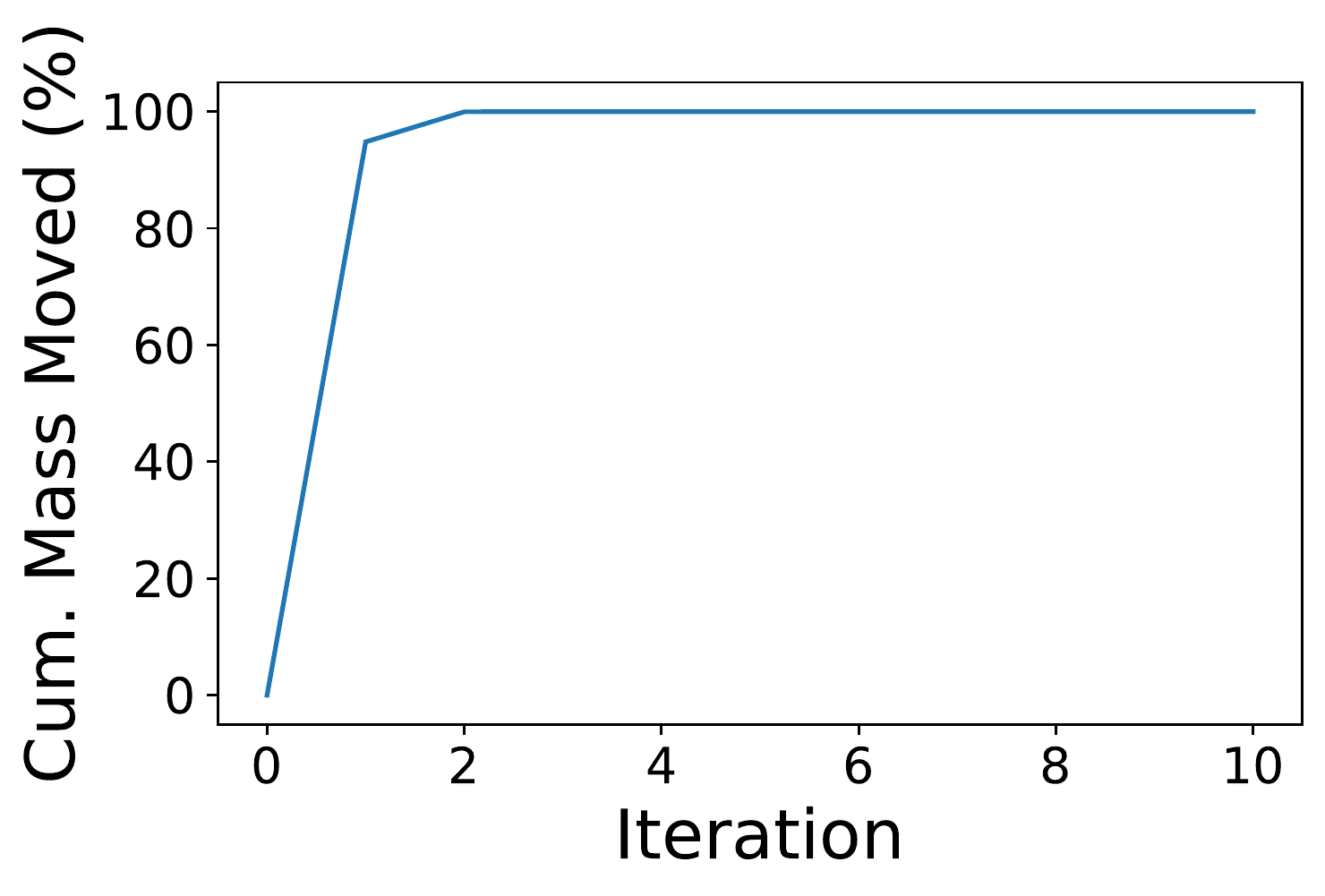}
        \caption{$\sigma=100$, $P=100$}
    \end{subfigure}
    \caption{Convergence of the mass-moving algorithm under random weight distributions. 
    In all settings, In all settings, the unpurified effect matrix is drawn from $N(0, \sigma I)$ of dimension $P$ while the weighting is drawn from a $P$-dimensional normal distribution $N(0, \sigma I)$. 
    Errorbars indicate the mean $\pm$ stddev. over 100 experiments. 
    In all settings, the algorithm converges in a small number of  iterations.\label{fig:convergence_random}}
\end{figure}

\end{document}